\documentclass[runningheads]{llncs}
\usepackage[T1]{fontenc}
\usepackage[english]{babel}

\usepackage{amsmath}
\usepackage{amssymb}

\usepackage{graphicx}
\usepackage{array}
\usepackage{booktabs}
\usepackage{multirow}
\usepackage{colortbl}

\usepackage{xcolor}
\usepackage[normalem]{ulem}

\usepackage{hyperref}

\usepackage[capitalize]{cleveref}

\input{preamble}

\begin{document}
\title{\lunargr: Geometry-to-Reflectance Learning \\for High-Fidelity Lunar BRDF Estimation}
%
%
\titlerunning{Geometry-to-Reflectance Learning for Lunar BRDF Estimation}
\author{
Clémentine Grethen\inst{1}\orcidID{0009-0009-3695-1717} \and
Nicolas Menga\inst{2} \and
Roland Brochard\inst{2} \and
Géraldine Morin\inst{1}\orcidID{0000-0003-0925-3277} \and
Simone Gasparini\inst{1}\orcidID{0000-0001-8239-8005} \and
Jérémy  Lebreton\inst{2}\orcidID{0000-0003-1476-5963} \and
Manuel Sanchez-Gestido\inst{3}\orcidID{0009-0003-0158-4300}
}

\authorrunning{C. Grethen et al.}

\institute{
IRIT, University of Toulouse, France \and
Airbus Defence and Space, Toulouse, France \and
European Space Agency (ESA) ESTEC, Noordwijk, The Netherlands
}

\maketitle              
\setcounter{footnote}{0}

\begin{figure}
    \centering
    \includegraphics[width=1\linewidth]{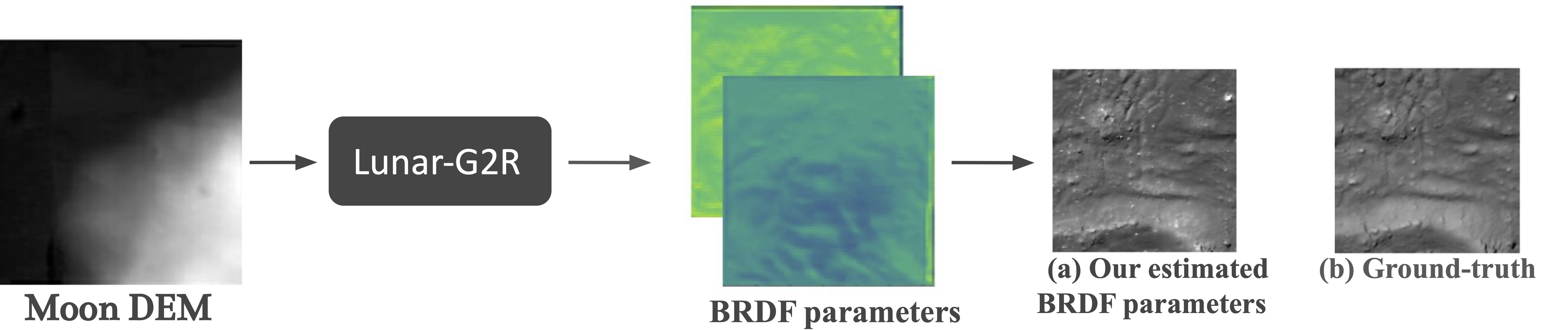}
    \caption{\lunargr  estimates a spatially-varying parameterized BRDF of the lunar surface given the geometry. This estimated appearance allows for a realistic rendering of the moon surface (a), which closely matches the ground-truth observation (b).}
    \label{fig:intro}
\end{figure}



\begin{abstract}
We address the problem of estimating realistic, spatially varying reflectance for complex planetary surfaces such as the lunar regolith, which is critical for high-fidelity rendering and vision-based navigation. 
Existing lunar rendering pipelines rely on simplified or spatially uniform BRDF models whose parameters are difficult to estimate and fail to capture local reflectance variations, limiting photometric realism. 
We propose \lunargr, a geometry-to-reflectance learning framework that predicts spatially varying BRDF parameters directly from a lunar digital elevation model (DEM), without requiring multi-view imagery, controlled illumination, or dedicated reflectance-capture hardware at inference time. 
The method leverages a U-Net trained with differentiable rendering to minimize photometric discrepancies between real orbital images and physically based renderings under known viewing and illumination geometry. 
Experiments on a geographically held-out region of the Tycho crater show that our approach reduces photometric error by \SI{38}{\percent} compared to a state-of-the-art baseline, while achieving higher PSNR and SSIM and improved perceptual similarity, capturing fine-scale reflectance variations absent from spatially uniform models. 
To our knowledge, this is the first method to infer a spatially varying reflectance model directly from terrain geometry.

\end{abstract}

\section{Introduction}
\label{sec:introduction}
Achieving physically realistic and predictive rendering requires accurate modeling of three fundamental components: surface geometry, material reflectance properties, and illumination.
Material reflectance is classically modeled by a Bidirectional Reflectance Distribution Function (BRDF), which governs how a surface scatters light and ultimately determines its appearance.
Estimating BRDF parameters that enable reliable re-rendering under arbitrary viewpoints and illumination remains a central and ill-posed challenge in computer vision and computer graphics~\cite{Guarnera2016}.

Accurate BRDF modeling is critical for space applications.
The design and validation of Vision-Based Navigation (VBN) algorithms rely heavily on synthetic imagery, whose realism depends on the underlying reflectance model.
The need for an accurate lunar regolith reflectance model is particularly acute for lunar missions, where precise photometric behavior is essential for robust perception and navigation.
Analytical BRDF models such as Hapke~\cite{hapke1993theory} provide a physically grounded description of lunar reflectance; however, in practice, only a subset of their parameters can be estimated at high spatial resolution.
The remaining parameters are typically fixed or assumed spatially uniform, preventing current simulations from capturing the fine-scale reflectance variations observed in real lunar imagery.
This motivates the need to estimate a spatially varying BRDF (SVBRDF) at a finer and more local scale.

Classical BRDF estimation methods aim to sample a wide range of illumination and viewing directions to capture surface reflectance~\cite{Guarnera2016,Kavoosighafi2024}.
They typically rely on specialized acquisition setups, such as gonioreflectometers, or on controlled capture protocols involving systematic and controlled variations of lighting and viewpoint.
Although accurate, these approaches are costly, difficult to scale, and incompatible with planetary settings, where multi-view or multi-illumination observations of the same surface are unavailable and specialized acquisition setups cannot be deployed in Space.

The lunar terrain presents a distinctive scenario where the surface geometry is well characterized and stored as Digital Elevation Models (DEMs)~\cite{lrodata,Henriksen2016}, while reflectance is challenging to model accurately due to its spatial variability.
This peculiar setting suggests the opportunity to infer reflectance properties directly from terrain geometry, provided that a physically based and differentiable image formation model is available.
The main challenge lies in modeling the interaction between surface morphology, illumination, and material composition under realistic imaging conditions. 
Recent advances integrate differentiable rendering directly into neural pipelines, enabling reflectance estimation by minimizing photometric errors between rendered and ground-truth images~\cite{Kavoosighafi2024}. 
In this work, we follow this paradigm and leverage SurRender~\cite{brochard2018scientific} — a physically accurate and differentiable ray-tracing engine tailored to space applications—to supervise reflectance learning using real lunar images.

We introduce \lunargr, a neural framework that predicts dense, spatially varying BRDF parameter maps at each point of a lunar DEM.
The framework is built around a U-Net architecture and trained using a differentiable rendering formulation, where images rendered from the predicted per-pixel BRDF parameters are compared to real orbital observations.
By enabling physically based renderings at higher spatial resolution and fidelity than standard analytical models such as Hapke, our approach significantly improves the realism of simulated lunar imagery.
This design allows reflectance properties to be learned directly from terrain geometry and demonstrates that estimating reflectance from geometry alone is feasible. 
This suggests that the same strategy could be extended to other lunar regions or planetary terrains whenever both elevation models and real images are available. 
To our knowledge, this is the first approach that learns a spatially varying BRDF directly from surface geometry at DEM resolution (\SI{5}{\meter\per\px}), reducing photometric rendering error by \SI{38}{\percent} on average compared to a Hapke-based baseline.


\textbf{The contributions of this work are twofold:} 
(i) We introduce \lunargr, a U-Net–based neural architecture that predicts polynomial SVBRDF parameter maps from high-resolution lunar DEM tiles.
(ii) We develop a complete learning pipeline that (a) constructs a large-scale training dataset of over \num{80000} DEM–image pairs (publicly released \footnote{Data and code: \url{https://clementinegrethen.github.io/publications/Lunar-G2R}}) by pairing real lunar images with corresponding DEMs and acquisition metadata through a dedicated sampling strategy ensuring photometric diversity and BRDF observability (over \num{80000} DEM–image pairs), and (b) integrates the network into a differentiable rendering framework to provide end-to-end, physically consistent supervision. 



The paper is organized as follows.
\Cref{sec:related} reviews prior work on BRDF and SVBRDF modeling and estimation, as well as existing approaches to lunar reflectance modeling and its use in planetary image simulation.
\Cref{sec:method} presents the \lunargr framework, including the BRDF parameterization and network architecture.
\Cref{sec:pipeline} details the training pipeline, covering dataset construction and differentiable rendering–based loss.
\Cref{sec:evaluation} reports the evaluation protocol and experimental results.
Finally, \Cref{sec:conclusion} discusses limitations and future directions.

\section{Related work}
\label{sec:related}
This section reviews prior work on BRDF and SVBRDF representations and estimation, before focusing on reflectance modeling and simulation in the lunar context.
\paragraph{\textbf{BRDF and SVBRDF}}
A joint representation of surface geometry and material appearance is essential for numerous vision and graphics pipelines, including view synthesis, relighting~\cite{Hofherr2025Neural,Guarnera2016}, and high-fidelity synthetic data generation for perception and simulation~\cite{Mumuni_2024}.
These applications rely on accurate models describing the directional reflectance behaviour of a material.
This is commonly expressed through Bidirectional Reflectance Distribution Functions (BRDFs) and their spatially varying extension (SVBRDFs) that enable reflectance parameters to change across the surface~\cite{Nicodemus1965}.
From a physical point of view, surface appearance is governed by the integration of the (SV)BRDF and the incoming illumination~\cite{Kajiya1986Rendering}.
Popular BRDF models include empirical formulations such as Lambert~\cite{Lambert1760} or Phong~\cite{Phong1975Illumination}, and physically-based models like the Torrance–Sparrow microfacet model~\cite{Cook1982AReflectance}. 
Measured data-based BRDFs are commonly represented either as dense lookup tables, as in the MERL database~\cite{Matusik2003}, or through compact parametric models fitted to such measurements (e.g. Disney BRDF~\cite{Burley2012PhysicallyBasedSA}).
More recently, neural BRDF representations~\cite{Sztrajman2021,Kavoosighafi2024} offer a compact alternative, reconstructing full BRDFs from sparse measurements and enabling BRDF editing and interpolation in a latent space~\cite{gokbudak2024hypernetworksgeneralizablebrdfrepresentation,Zheng2021NeuralProcessBRDF}. 

\paragraph{\textbf{BRDF and SVBRDF estimation}}
Traditional BRDF estimation has mostly relied on controlled acquisition setups designed to densely sample reflectance ~\cite{Guarnera2016,Kavoosighafi2024}. Typically, Gonioreflectometers sample light and camera directions to measure appearance over a large set of incident and outgoing directions, as in the MERL database acquisition pipeline. 
These systems deliver highly accurate BRDF measurements, but at the cost of complex hardware and thousands of samples per material. 
Extending this to spatially varying reflectance (SVBRDF) requires sampling reflectance across both space and directions, typically using specialized capture setups such as spherical gantries~\cite{Dana1999,Ma2023}.
Despite highly accurate measures, these acquisitions are costly and restricted to laboratory conditions.
To mitigate this, prior-based SVBRDF methods restrict reflectance to a low-dimensional representation — such as linear combinations or manifold models of BRDF basis — and recover spatially varying reflectance from a small number of flash-lit images~\cite{nam2018practical,dong2010manifold}.

A complementary strategy focuses instead on increasing the information content of each measurement.
Illumination multiplexing methods encode multiple lighting conditions into each image, improving acquisition efficiency and signal-to-noise ratio~\cite{Ghosh2009}.
Recent approaches further integrate deep learning to directly infer reflectance parameters and jointly optimize illumination patterns and reconstruction~\cite{kang2018efficient,ma2021free}.

Beyond these classical methods, inverse-rendering approaches jointly optimize geometry, materials, and illumination from multiple images by coupling differentiable rendering with optimization through neural scene representations.
Recent works based on neural Signed Distance Functions (SDFs), neural reflectance fields, and NeRF-style~\cite{mildenhall2020nerf} pipelines decompose appearance into geometry, BRDF/SVBRDF parameters, and lighting, enabling relighting and material editing from multi-view images~\cite{Zhang_2021,boss2021nerdneuralreflectancedecomposition,boss2021neuralpilneuralpreintegratedlighting,sarkar2023litnerf}.
Complementary to these pipelines, learning-based SVBRDF methods recover spatial maps of diffuse/specular albedo, roughness, and normals from one or a few photographs using CNNs trained on synthetic or measured materials~\cite{Deschaintre2018,Guo2021Highlight}; other image-based approaches, such as Material Palette~\cite{lopes2023materialpaletteextractionmaterials} segment an image into a small number of material regions and estimate a parametric SVBRDF for each of them.

Similar to these methods, our approach uses differentiable rendering to supervise reflectance estimation by backpropagating photometric errors between rendered and observed images.
However, it differs fundamentally in both the learned representation and the nature of data.
Rather than learning a fully implicit reflectance field or per-image material maps from multi-view or controlled acquisitions, \lunargr learns spatially varying BRDF parameters defined over terrain geometry to provide an explicit, analytic BRDF formulation. 
Thus, the result is a neural network that, given the terrain geometry, computes a (parameterized) BRDF map at the same resolution as the input geometry/DEM.
Moreover, the model is trained from a limited set of real orbital images paired with DEMs and acquisition metadata. This design makes \lunargr compatible with planetary constraints, where controlled lighting, repeated observations, and dense multi-view image collections are unavailable.

\paragraph{\textbf{Lunar Reflectance Modeling and Simulation}}
\label{moon}
In the lunar context, the Hapke model has been widely used to study surface photometry and is particularly representative of the dusty lunar regolith: an analytical BRDF model that expresses surface reflectance as an explicit function of the illumination and viewing geometry, \ie, the incidence, emission, and phase angles.
The model is parameterized by a set of six physically meaningful coefficients controlling the angular reflectance behavior, including the single-scattering albedo, the particle phase function, the macroscopic roughness, and opposition effects \cite{Sato2014}.
In particular, the Shadow Hiding Opposition Effect (SHOE)~\cite{Hapke1986} accounts for the brightness surge observed at very small phase angles due to the disappearance of mutual shadows between regolith particles. 
Albedo is therefore only one component of the Hapke reflectance model. 


Physically based reflectance models are essential for generating realistic lunar image simulations used to design and evaluate vision-based navigation (VBN) algorithms. 
Despite the availability of such models, several publicly available synthetic lunar datasets rely on non-physical or heuristic assumptions: for example, the LuSNAR dataset~\cite{liu2024lusnaralunarsegmentationnavigation} provides synthetic lunar imagery generated with the Unreal Engine, focusing on visual plausibility, but without explicit modeling of lunar surface reflectance.
In contrast, several physically grounded lunar datasets rely on planetary rendering engines to generate synthetic imagery under realistic imaging conditions.
Physics-based simulators such as Pangu~\cite{Parkes2004} and SurRender~\cite{brochard2018scientific,lebreton2022high} combine ray tracing, high-resolution DEMs, physically based illumination, and a Hapke-based reflectance model to simulate lunar images for perception and navigation studies. 
Recent datasets such as StereoLunar~\cite{grethen2025adaptingstereovisionobjects} build upon these tools to produce photometrically consistent stereo imagery with full geometric supervision.
Although the Hapke BRDF is, in principle, parameterized by six physically meaningful coefficients that could vary spatially, high-resolution maps are available only for albedo. 
As a result, the remaining Hapke parameters are fixed to constant default values over the entire surface, and spatial variability is largely limited to albedo maps. 
In practice, even this variability is strongly constrained by the coarse resolution of available albedo maps, such as the Clementine mosaic~\cite{Lee2009ClementineBasemap} (\SI{118}{\metre\per\px}), whose resolution is significantly lower than that of commonly used DEMs (typically \SI{\approx 5}{\metre\per\px}).
Consequently, despite the use of physically based rendering engines, the effective reflectance model reduces to a single BRDF shared across the surface, whose variations are dominated by illumination and viewing geometry rather than by local material or geological properties.
This prevents such simulations from reproducing the fine-scale brightness variations observed in real lunar imagery.

Thus, fixing the Hapke parametric coefficients turns the nominal SVBRDF into a single BRDF shared over the surface, preventing spatially varying reflectance.
This motivates the development of \lunargr, which estimates spatial maps of BRDF coefficients at each surface location, that is, providing a reflectance map at a resolution similar to the geometry/DEM resolution. We thus provide  an actual SVBRDF modeling that enables generating more realistic lunar image simulation than what is currently available.

\section {Our method: a spatially variant lunar BRDF estimation}
\label{sec:method}

\begin{figure}[!h]
    \centering
    \includegraphics[width=0.9\linewidth]{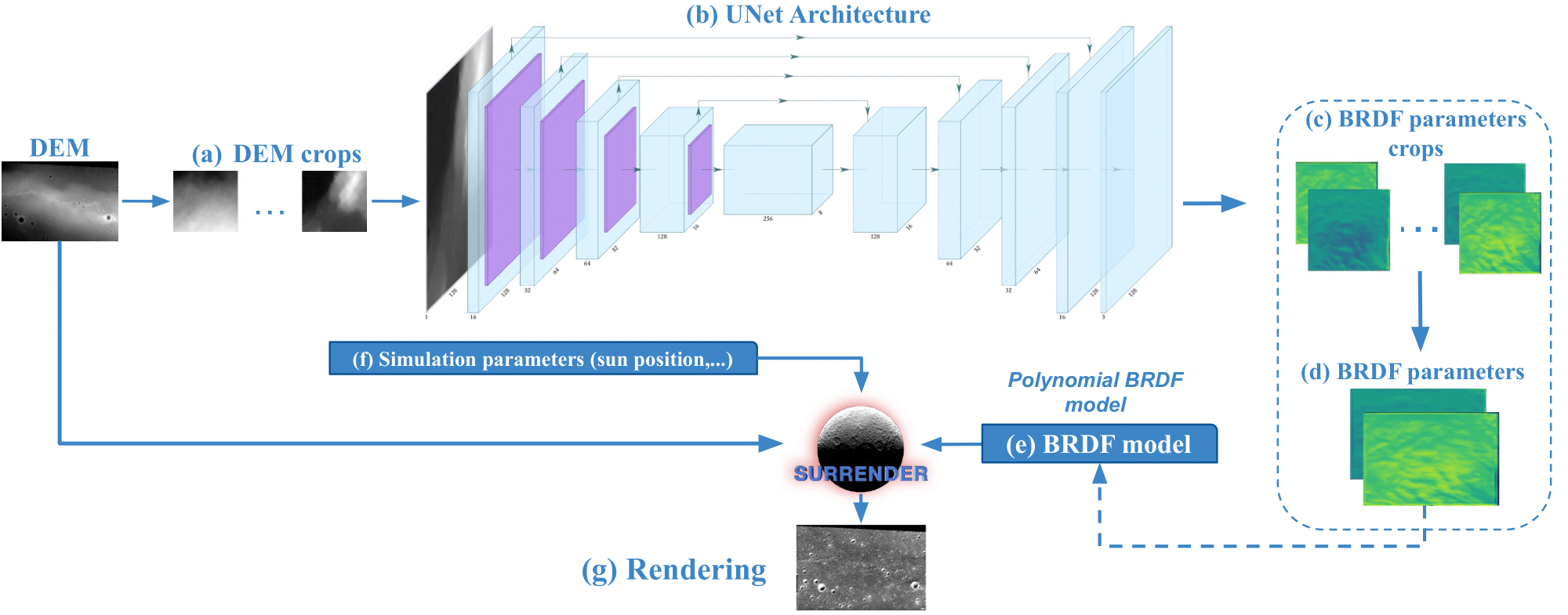}
    \caption{\lunargr inference pipeline for SVBRDF estimation from DEMs.}

    \label{fig:framework}
\end{figure} 
This section introduces \lunargr (see \cref{fig:framework}), a neural generator of SVBRDF parameter maps (see \cref{sssec:polynomial_brdf} for the choice of the SVBRDF model) corresponding to an input lunar DEM (\cref{sec:input}). 
Its architecture, detailed in \cref{ssec:network_architecture}, is based on U-Net and trained within a differentiable rendering framework using real Moon images.

\subsection{Input}
\label{sec:input}

Our approach is based on the assumption that material reflectance is locally correlated with surface morphology (crater rims, rocks, slopes, \etc), allowing reflectance properties to be inferred  directly from terrain geometry~\cite{Pieters2016-dm}.
While this hypothesis may not hold universally, we show that our approach produces synthetic images that more closely resemble real observations than those rendered using existing reflectance models, such as Hapke-based renderings, making the approach valuable even when the correlation is only partial.
Based on this assumption, the input of \lunargr consists solely of a digital elevation model (DEM), as shown in \cref{fig:framework}.
The DEM is first divided into fixed-size overlapping crops expressed in physical scale through the Ground Sample Distance (GSD) (\cref{fig:framework} (a)), \ie, the distance between two consecutive pixel centers on the ground.
This ensures that each patch corresponds to a consistent surface extent and preserves the geometric level of detail seen during training.
To ensure stable learning across heterogeneous terrain, DEM values undergo a global normalization (\cf \cref{sec:dem_normalization}), which removes sensitivity to absolute altitude differences between crops.
The model processes these patches in batches and reassembles the local predictions into a single georeferenced output, enabling BRDF estimation on arbitrarily large DEMs while maintaining spatial continuity.

\subsection{BRDF models}

\label{sssec:polynomial_brdf}

While the Hapke model, shown in \cref{moon}, has been extensively used for planetary photometry, its highly non-linear formulation makes it difficult to integrate into a stable and efficient gradient-based optimization pipeline.
Rather than attempting to invert the Hapke model directly, we consider a family of simpler and fully differentiable cosine-polynomial BRDF formulations, which preserve physical interpretability while remaining compatible with learning-based optimization.

All candidate BRDFs are expressed as low-order polynomials of angular quantities derived from the illumination and viewing geometry.
We primarily rely on the incidence angle $\theta_i$ and the phase angle $\theta_p$, which capture the dominant angular dependencies of reflectance for rough planetary surfaces.
To explore the trade-off between expressiveness and model complexity, we evaluate several polynomial BRDF formulations of increasing capacity, summarized in \cref{tab:brdf_models}.
In addition to models based on $(\theta_i,\theta_p)$, we also consider a polynomial formulation relying on the Rusinkiewicz \cite{Rusinkiewicz1998} half-vector parameterization, which re-expresses the incident and outgoing directions using the half-angle $\theta_h$, the difference angle $\theta_d$, and their associated azimuthal angles $(\varphi_h,\varphi_d)$, thereby explicitly modeling azimuthal reflectance effects.
Further details on the formulation are given in Section 2 of the supplementary material.

For all models, the polynomial coefficients $a$, $b$ (and $c$, $d$ when applicable) are the BRDF parameters predicted by \lunargr.
Since these coefficients are not global values but are estimated independently at each spatial location $(x,y)$ of the DEM, the BRDF becomes spatially varying, \ie,
\begin{equation}
(x,y) \mapsto \big(a(x,y),\, b(x,y),\, c(x,y),\, d(x,y)\big).
\end{equation}

In other words, the SVBRDF is entirely encoded by the spatial variation of the polynomial parameters over the surface geometry.
This formulation enables per-pixel BRDF estimation and relaxes the assumption of spatially uniform reflectance commonly adopted in lunar rendering pipelines.
\begin{table}[t]
\centering
\caption{Cosine-polynomial BRDF candidate models evaluated in this work.}
\label{tab:brdf_models}
\begin{tabular}{l c l}
\toprule
\textbf{Model} & \textbf{\# params} & \textbf{BRDF formulation} \\
\midrule
(M1) 
& 2
& $a\cos(\theta_i) + b\cos(\theta_p)$ \\

(M2) 
& 3
& $a\cos(\theta_i) + b\cos(\theta_p) + c$ \\

(M3) 
& 4
& $a\cos(\theta_i) + b\cos(\theta_p)
   + c\cos(\theta_i)\cos(\theta_p) + d$ \\

(M4)  
& 4
& $a\cos(\theta_i) + b\cos(\theta_p)
   + c\cos^2(\theta_i) + d$ \\

(M5) 
& 4
& $a\cos(\theta_i) + b\cos(\theta_p)
   + c\cos^2(\theta_p) + d$ \\

(M6) 
& 4
& $a\cos(\theta_h) + b\cos(\theta_d)
   + c\cos(\varphi_h) + d\cos(\varphi_d)$ \\
\bottomrule
\end{tabular}
\end{table}

\subsection{Network Architecture}
\label{ssec:network_architecture}
Our neural network predicts, for each $(x,y)$ from the geometric input (DEM), the SVBRDF parameters $(a,b)$, $(a,b,c)$, or $(a,b,c,d)$ defined in \cref{tab:brdf_models}, depending on the selected polynomial BRDF model.
Since both input and output are spatially structured, we adopt a Convolutional Neural Network (CNN) architecture.
In particular, a U-Net~\cite{ronneberger2015unetconvolutionalnetworksbiomedical} is employed (\cref{fig:framework} (b)), which offers a better trade-off between accuracy and complexity for this kind of dense prediction task.
Its skip connections enable the network to combine fine-scale geometric features with larger contextual information, while maintaining spatial resolution—key for accurate SVBRDF estimation.
The network U-Net is a four–stage U-Net ($\approx$ 2M parameters), with max–pooling for downsampling and nearest–neighbor upsampling followed by convolution.
Each block uses Batch Normalization and a Leaky ReLU activation, which stabilizes optimization while preserving positive BRDF parameter responses.
This design preserves local detail while covering the larger spatial context needed for reflectance estimation.
During inference, the U-Net predicts BRDF maps for each DEM crop (\cref{fig:framework} (c)).
The local predictions are then reassembled into their original spatial positions so that every pixel of the DEM is assigned its own set of BRDF coefficients.
The final output is therefore a three-channel SVBRDF map, where each channel stores one of the coefficients $a(x,y)$, $b(x,y)$, $c(x,y)$, and $d(x,y)$ (\cref{fig:framework} (d)).
The resulting georeferenced BRDF parameter map, together with our polynomial BRDF model (\cref{fig:framework} (e)), can be directly used for image rendering. Combined with the input DEM and any user-defined simulation setup (camera geometry, illumination conditions, \etc) (\cref{fig:framework} (f)), it enables realistic synthesis using SurRender (\cref{fig:framework} (g)) or any compatible rendering engine.

\section{Training pipeline} 
\label{sec:pipeline}
For each training sample, our method uses a high-resolution DEM and a real lunar image of the same region (used as ground truth), together with the acquisition metadata: camera pose, illumination geometry, and projection parameters.
These inputs enable us to reproduce the real imaging conditions in a differentiable rendering framework.
After predicting the BRDF from the DEM, we render a synthetic image under the same viewing and lighting configuration, and back-propagate the photometric error \wrt the ground-truth image through the renderer.
This enables learning the coupling between terrain, illumination, and reflectance without controlled photometric setups or multi-view acquisition.
The training pipeline is structured in two components: (i) dataset construction and (ii) differentiable rendering.

\subsection{Dataset design}
\label{dataset}
For our dataset, we generate paired samples associating local lunar terrain geometry with
corresponding appearance observations.
Each sample consists of a DEM crop of size \(128 \times 128\) pixels at a GSD
of \SI{5}{\meter\per\px}, corresponding to an area of approximately \SI{0.4}{\kilo\meter\squared},
extracted from a large DEM that covers the Tycho crater.
For each DEM crop, a real Moon image acquired by the Lunar Reconnaissance Orbiter (LRO)~\cite{lrodata} covering the corresponding terrain region is selected; it is then orthorectified onto the local DEM, and cropped to exactly match the spatial extent of the DEM patch, producing a ground-truth appearance image aligned with the topography.
For each pair, we store the acquisition metadata associated with the LRO image, including the camera pose, Sun illumination direction (the sole light source in the lunar environment), field of view, and geographic footprint. The source LRO images have native resolutions ranging from \qtyrange{0.5}{2}{\meter\per\px}.

\begin{figure}
    \centering
    \includegraphics[width=0.8\linewidth]{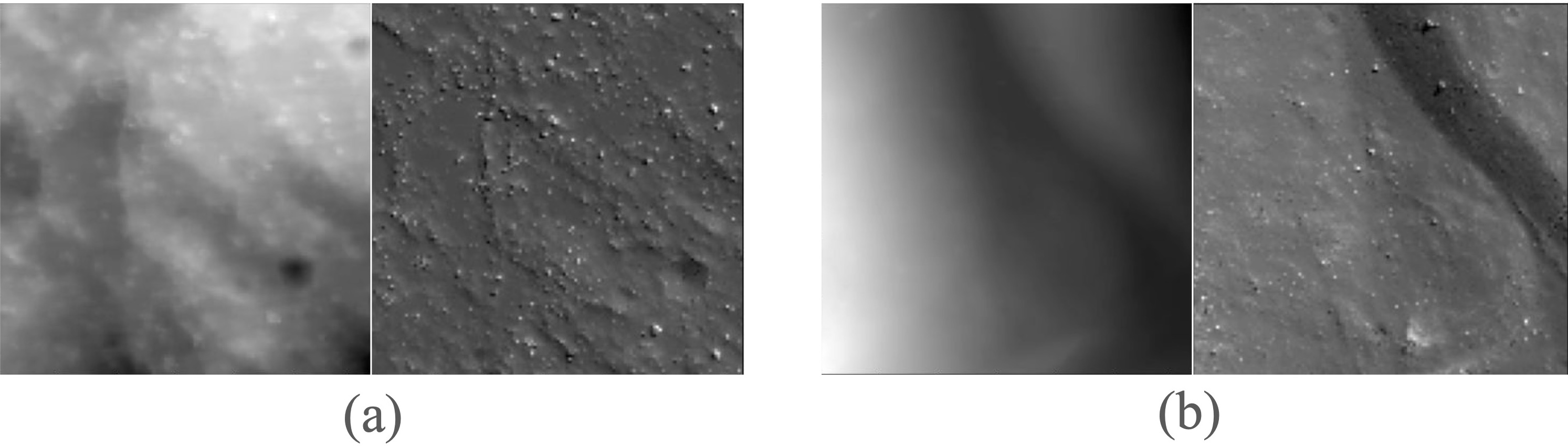}
    
    \caption{Examples of DEM--image pairs from the dataset: cropped DEM patches (left) and corresponding orthorectified lunar images (right).}
    
    \label{fig:exdata}
\end{figure}

In total, the final dataset contains \textbf{83,614} DEM–image ($128\times128$ px) pairs with metadata, partitioned into \textbf{66,662} training, \textbf{8,615} validation, and \textbf{8,337} test samples (see \cref{fig:exdata} for examples).
To avoid spatial leakage, dataset splits are performed geographically after pair generation: the global DEM is divided into tiles, and all pairs whose centers fall within the same tile are assigned to the same split. 
Additional details on dataset construction and preprocessing are provided in Section 1 of the supplementary material.

\subsubsection{DEM normalization.}
\label{sec:dem_normalization}
Although the global lunar relief spans elevations from \SIrange{-5.6}{7.5}{\kilo\metre}~\cite{LRO_Coordinate_System_2008}, each DEM crop covers only a small spatial region (\SI{\approx0.4}{\kilo\metre\squared}), and its height range varies significantly depending on local morphology.
As a result, normalizing each crop independently would produce heterogeneous representations and hinder training. 
Instead, we center each crop by subtracting its mean elevation, making the model agnostic to local absolute altitude, and scale elevations using a dataset-wide standard deviation to ensure consistent variation magnitudes across samples.
This normalization preserves relative terrain variations while improving training stability and preventing the network from encoding absolute elevation.

\subsection{Differentiable rendering and loss formulation}
\begin{figure}
    \centering
    \includegraphics[width=0.9\linewidth]{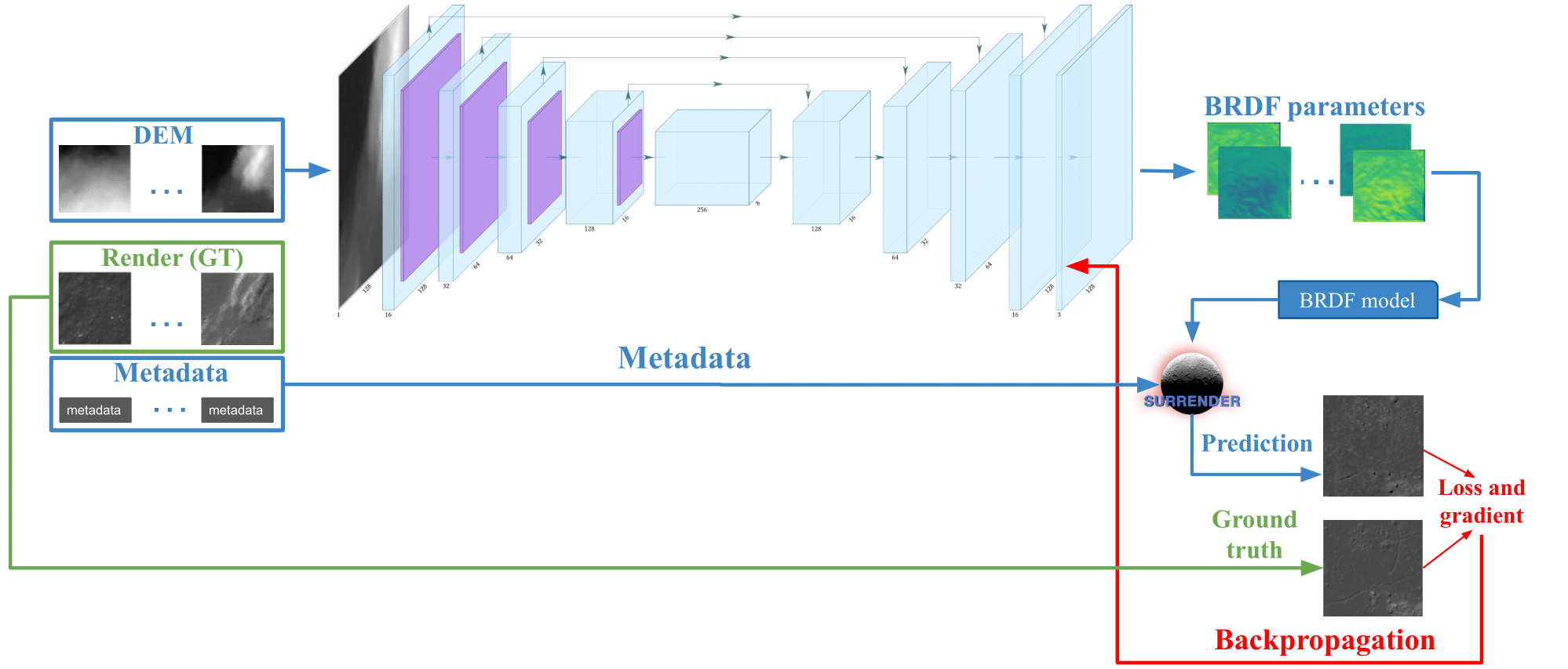}
    \caption{Training pipeline with differentiable rendering in \lunargr.}
    \label{fig:train-pipeline}
\end{figure}
Each normalized DEM crop is fed into the U-Net, which predicts the local BRDF coefficients of the polynomial reflectance model (\cref{tab:brdf_models}).
For each training pair, the associated acquisition metadata (camera pose, Sun direction, and field of view) are used to condition a SurRender layer that performs physically-based rendering of the DEM crop using the predicted BRDF parameters (\cref{fig:train-pipeline}).
The resulting rendered image is then compared to the corresponding projected ground-truth appearance image, and the photometric discrepancy is measured using a Mean Squared Error (MSE) loss. 
Since SurRender provides analytical gradients \wrt the BRDF parameters, this loss can be back-propagated through the rendering process and into the U-Net, enabling end-to-end learning of the reflectance model.

 The network is trained for 20 epochs with Adam (lr $10^{-4}$, batch size $64$) on a single NVIDIA RTX~A4500 (\SI{16}{\giga\byte}), for a total training time of $\sim$36~h.
 
\section{Evaluation}
\label{sec:evaluation}
In the following, we evaluate the proposed method using quantitative metrics. 
We first compare the different candidate BRDF models to identify an appropriate reflectance formulation. 
We then report photometric evaluation results using the selected BRDF model. 
An additional analysis based on feature-matching consistency is reported in Section 5 of the supplementary material.

\subsection{BRDF model comparison}
\label{sec:brdf_evaluation}

\begin{figure}
    \centering
    \includegraphics[width=0.9\linewidth]{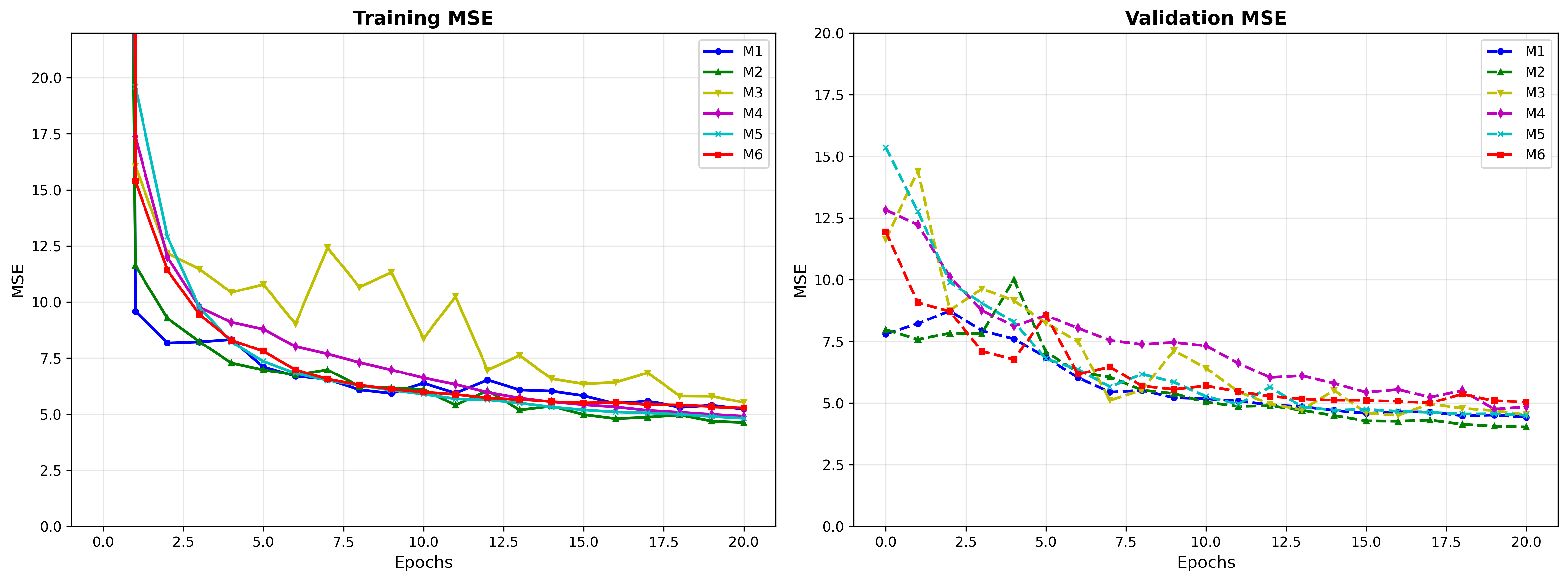}
\caption{Training (left) and validation (right) loss curves across epochs for the polynomial BRDF models.}
    \label{fig:brdf_training_curves}
\end{figure}

\cref{fig:brdf_training_curves} compares the training and validation MSE obtained with the polynomial BRDF models listed in \cref{tab:brdf_models}.
All formulations converge rapidly, confirming that low-order cosine polynomials capture the dominant angular reflectance effects, despite less stable optimization for (M3). Among them, the 3-parameter model (M2) consistently achieves the lowest validation error. In contrast, more expressive 4-parameter variants (M4, M5, M6) do not yield additional gains and exhibit higher validation errors. This indicates that under the limited angular distribution, higher-order terms are largely redundant and add no new information. At convergence, the 3-parameter model (M2) reaches a validation MSE of $4.06$, compared to $4.51$ for (M1) and $5.34$ for (M6). Based on these results, we retain the  model (M2) for the remainder of this work.

\subsection{Photometric Evaluation}
\label{photometric}

\begin{table}[h]
\centering
\begin{tabular}{lcccc}
\toprule
 & \textbf{MSE ↓} & \textbf{PSNR ↑} & \textbf{SSIM ↑} & \textbf{LPIPS ↓} \\
\midrule
\textbf{Ours} & \textbf{4.81} & \textbf{23.14} & \textbf{0.520} & \textbf{0.25} \\
Hapke (normalized) & 7.71 & 21.02 & 0.411 & 0.32 \\
\bottomrule
\end{tabular}
 \caption{Quantitative comparison between our learned BRDF and a normalized Hapke baseline. Lower MSE and LPIPS, and higher PSNR and SSIM indicate improved photometric fidelity to real images.}
\label{tab:metrics}
\end{table}

In the following, all photometric evaluations are conducted using the selected  BRDF model (M2).
We evaluate our framework in a zero-shot setting on the held-out geographical test split described in \cref{dataset}.
The quality of the predicted BRDF is assessed by comparing renderings produced with the learned parameters against the ground-truth appearance provided in the dataset, using the same acquisition metadata (Sun position and camera pose) associated with each sample.
As a baseline, we use a Hapke-based model rendered with the same SurRender configuration and fixed (default) Hapke parameters over the entire surface (\cref{sec:related}).
Since the standard Hapke model does not include spatially varying albedo and its absolute photometric scale differs from the ground truth, a direct comparison would be biased and unfair.
Instead, we use a normalized Hapke variant, in which each rendered output is rescaled per sample to match the dynamic range of the ground–truth image.
This preserves Hapke's photometric behaviour while removing the radiometric offset, resulting in a more realistic and fair baseline for comparison.
All images (ours, Hapke, and the ground truth) are rendered under the  same acquisition metadata settings.
To quantify the photometric accuracy, we report four complementary metrics: Mean Squared Error (MSE), a pixel-wise fidelity measure; PSNR and SSIM~\cite{Wang2004SSIM}, which capture luminance and structural similarity; and LPIPS~\cite{Zhang2018LPIPS}, a perceptual distance based on deep features.
A summary of the results is given in \cref{tab:metrics}.

\begin{figure}[t]
    \centering
    \includegraphics[width=0.8\linewidth]{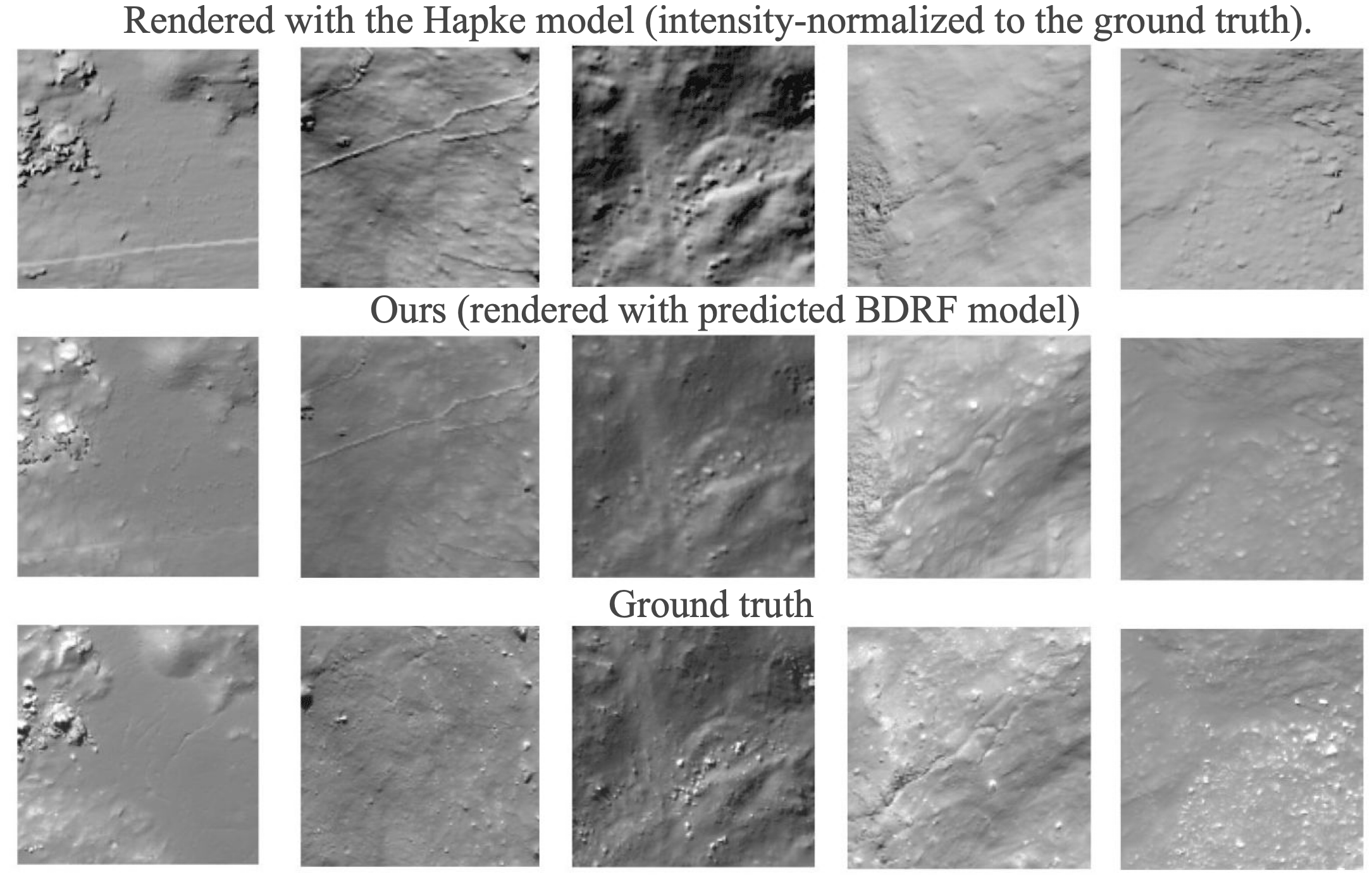}
    \caption{Qualitative comparison across lunar regions: each column shows a different area with varying geometry and reflectance under real acquisition conditions. Additional examples are in Section 3 of the supplementary material.}
    \label{fig:metrics}
\end{figure}

Our model achieves a substantial improvement in photometric fidelity compared to the Hapke baseline.
As shown in \cref{tab:metrics}, the learned BRDF reduces the MSE by \SI{38}{\percent} relative to the normalized Hapke model and yields higher PSNR and SSIM scores, indicating a closer match to the ground-truth appearance.
In particular, the improvement in SSIM ($0.520$ \vs $0.411$) reflects a significantly better preservation of local structures and contrast, which is consistent with the qualitative observations.
This trend is further supported by a lower LPIPS score, indicating that the renderings produced with the learned SVBRDF are perceptually closer to real observations.
As illustrated in \cref{fig:metrics}, our method better reproduces fine-scale brightness variations on slopes and rough terrain, whereas the Hapke baseline remains overly smooth and spatially uniform.
Overall, these results confirm that learning spatially varying reflectance parameters from DEMs leads to more realistic image synthesis than relying on a fixed analytical photometric model. Building on these findings, we release an enhanced version of the StereoLunar\cite{grethen2025adaptingstereovisionobjects} dataset, in which synthetic stereo pairs are rendered using the estimated SVBRDFs instead of a fixed Hapke model, enabling more photorealistic simulations for downstream lunar 3D vision tasks.
Experiments on generalization to unseen viewpoints are provided in Section 4 of the supplementary material.

\section{Conclusion}
\label{sec:conclusion}
We introduced \lunargr, a framework that learns spatially varying BRDF parameters from lunar DEMs via differentiable rendering. 
Using our inferred SVBRDF predictions for simulating images significantly improves photometric accuracy and structural consistency over a Hapke baseline, while capturing fine-scale reflectance variations and producing more realistic lunar renderings.
Despite these promising results, the model remains limited by the characteristics of the training dataset, which predominantly contains near-nadir (i.e., viewing direction close to the surface normal) observations (\SI{70}{\percent} nadir and \SI{30}{\percent} oblique), covers only the Tycho crater region, and inherits artifacts from the underlying DEM. 
These factors constrain the angular and geological diversity of the learned reflectance and partly explain the residual discrepancies between simulated and real images.
Nevertheless, the method generalizes reasonably well to moderately oblique views and demonstrates that estimating reflectance from elevation data alone is feasible in practice.
More broadly, \lunargr defines a geometry-to-reflectance learning paradigm that is not specific to the Moon and could extend to other planetary bodies (e.g., Mars)---or more generally to any setting where elevation models and imagery are available, and the illumination/viewing conditions are known or can be reliably estimated.
Future work will focus on extending dataset diversity—particularly in terms of illumination, viewing geometry, and terrain types—and on exploring richer reflectance parameterizations to improve BRDF generalization and photorealistic image simulation.


{
\paragraph{Acknowledgements} This work was carried out with the support of the European Space Agency (ESA) under contract n°4000140461/23/NL/GLC/my and previously under ESA contract 4000145741/24/D/MB.
We thank Philippe Nonin from Airbus Pixel Factory, who produced the DEM used in the study.}

%
%
%
%
\clearpage

\definecolor{pastelgreen}{RGB}{220,245,220}
\definecolor{pastelblue}{RGB}{220,230,250}
\definecolor{pastelpink}{RGB}{250,220,230}

\title{\lunargr: Geometry-to-Reflectance Learning \\for High-Fidelity Lunar BRDF Estimation
(Supplementary material)}
\titlerunning{Geometry-to-Reflectance Learning for Lunar BRDF Estimation}
%
%
\author{
Clémentine Grethen\inst{1}\orcidID{0009-0009-3695-1717} \and
Nicolas Menga\inst{2} \and
Roland Brochard\inst{2} \and
Géraldine Morin\inst{1}\orcidID{0000-0003-0925-3277} \and
Simone Gasparini\inst{1}\orcidID{0000-0001-8239-8005} \and
Jérémy  Lebreton\inst{2}\orcidID{0000-0003-1476-5963} \and
Manuel Sanchez-Gestido\inst{3}\orcidID{0009-0003-0158-4300}
}

\authorrunning{C. Grethen et al.}

\institute{
IRIT, University of Toulouse, France \and
Airbus Defence and Space, Toulouse, France \and
ESA ESTEC, Noordwijk, The Netherlands
}

\maketitle              

\section{More details on the dataset design}
\label{dataset}
\begin{figure}[t]
    \centering
    \includegraphics[width=1\linewidth]{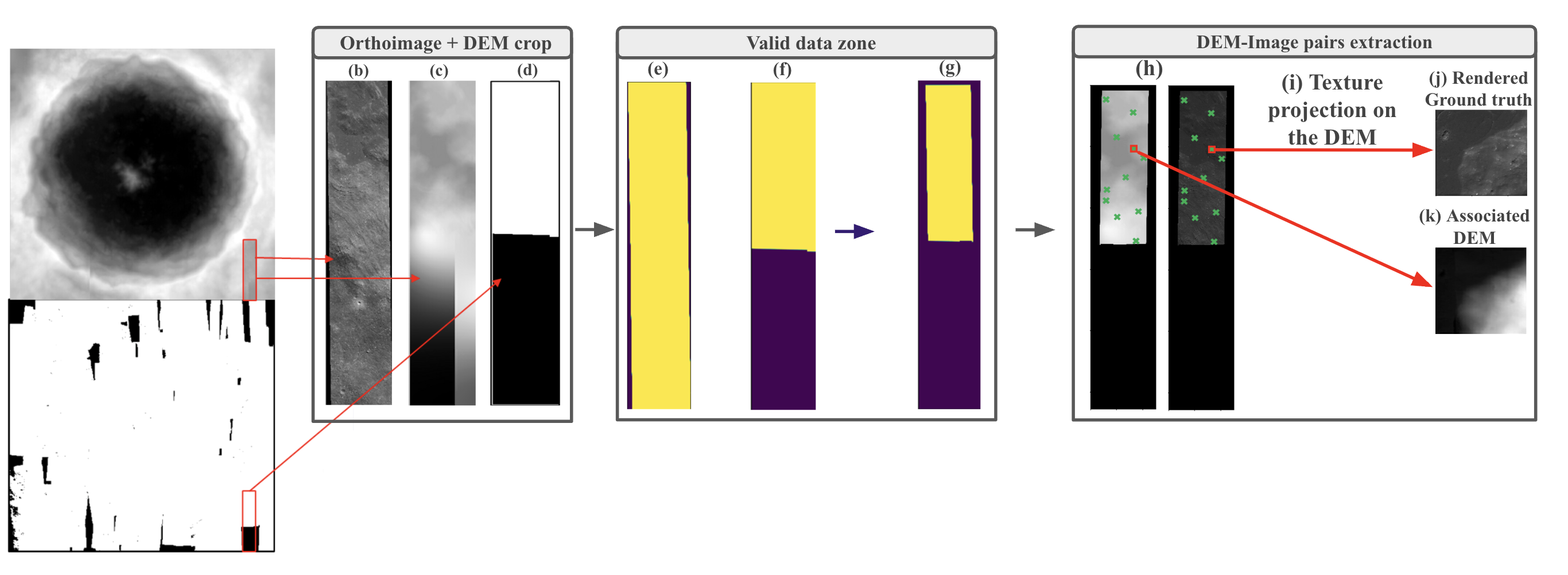}
    \caption{Dataset design pipeline}
    \label{fig:dataste-pipeline}
\end{figure}
This section provides additional details on the construction of the training dataset used in our study.

We train and evaluate our model on a dataset covering the Tycho crater~\cite{Krger2016}, whose size is $95 \times 90\,\mathrm{km}$  and elevation range between $-3570$ and $1856\,\mathrm{m}$.
The terrain elevation is provided by a high-resolution DEM produced with Airbus Pixel Factory%
\footnote{\url{https://space-solutions.airbus.com/imagery/imagery-processing/}}
from stereo satellite imagery, originally at \SI{1}{\meter\per\px} GSD. 
Pixel Factory is an industrial geo-production system developed by Airbus Sophia Antipolis, containing a state-of-the-art digital processing chain to produce advanced 2D and 3D mapping products of unparalleled quality. The DEM is downsampled by a factor of $5$ to obtain a cleaner (small artifacts are filtered) and more robust input (\cref{fig:dataste-pipeline}(a)). 

A set of $667$ ortho--rectified lunar images extracted from the LRO mission portal 
\cite{Henriksen2016,lrodata} provides the appearance data 
(\cref{fig:dataste-pipeline}(b)). 
The images exhibit native resolutions ranging from \qtyrange{0.5}{2}{\meter\per\px} and sizes between 
$5000\times 25000$ and $5000\times 50000$ pixels (\qtyrange{31}{1000}{\kilo\meter\squared}), excluding no-data regions. 
For each orthoimage, we extract the DEM over the same geographic footprint 
(\cref{fig:dataste-pipeline}(c)) and compute validity masks for both sources 
(\cref{fig:dataste-pipeline}(e)--(f)).

We aim to sample fixed-size local patches, or \emph{crops}, from the DEM and the appearance data; each crop corresponds to the input region processed by the neural network. 
To guarantee that a full crop lies within valid data, the merged mask is eroded 
(\cref{fig:dataste-pipeline}(g)) using a kernel size equal to half the physical crop footprint (\ie, $(s \cdot \mathrm{GSD})/2$, with $s$ the crop size in pixels). 
This defines the set of \emph{valid positions}, \ie, pixels that can safely serve as centers of crops whose full spatial extent contains only valid DEM and appearance data. 
Because orthoimages cover highly variable areas, the number of crops extracted from each one is  proportional to its valid surface, ensuring a homogeneous dataset over the Tycho region. 
With a crop size of $128\times128$ pixels at a target \SI{5}{\meter\per\px}, each crop covers \SI{\approx0.4}{\kilo\meter\squared}. 
Using these valid positions, we randomly select crop centers (\cref{fig:dataste-pipeline}(h)) and extract the corresponding DEM patches.
The associated LRO metadata (camera pose, Sun direction, field of view, and geographic footprint) is then used to geometrically project the orthoimage onto each local DEM patch (\cref{fig:dataste-pipeline}(i)), producing an appearance image aligned with the topography (\cref{fig:dataste-pipeline}(j)).
This projected image crop serves as the ground truth for the DEM patch.
Since the orthoimages strongly overlap, we prevent spatial leakage by applying a geographical split after generating all DEM–image pairs: the global DEM is divided into tiles, and all pairs whose centers fall within the same tile are assigned together to the train, validation, or test set.

In total, the dataset contains \textbf{\num{83614}} DEM--image pairs with metadata, partitioned into \textbf{\num{66662}} training,\textbf{ \num{8615}} validation, and \textbf{\num{8337}} test samples, providing homogeneous coverage of the Tycho crater.

\section{BRDF Model Comparison}
In this section, we provide further details on the angular representation used for BRDF parametrization, including the simple formulation (M1 to M5) and the Rusinkiewicz \cite{Rusinkiewicz1998} parameterization.

\begin{figure}
    \centering
    \includegraphics[width=1\linewidth]{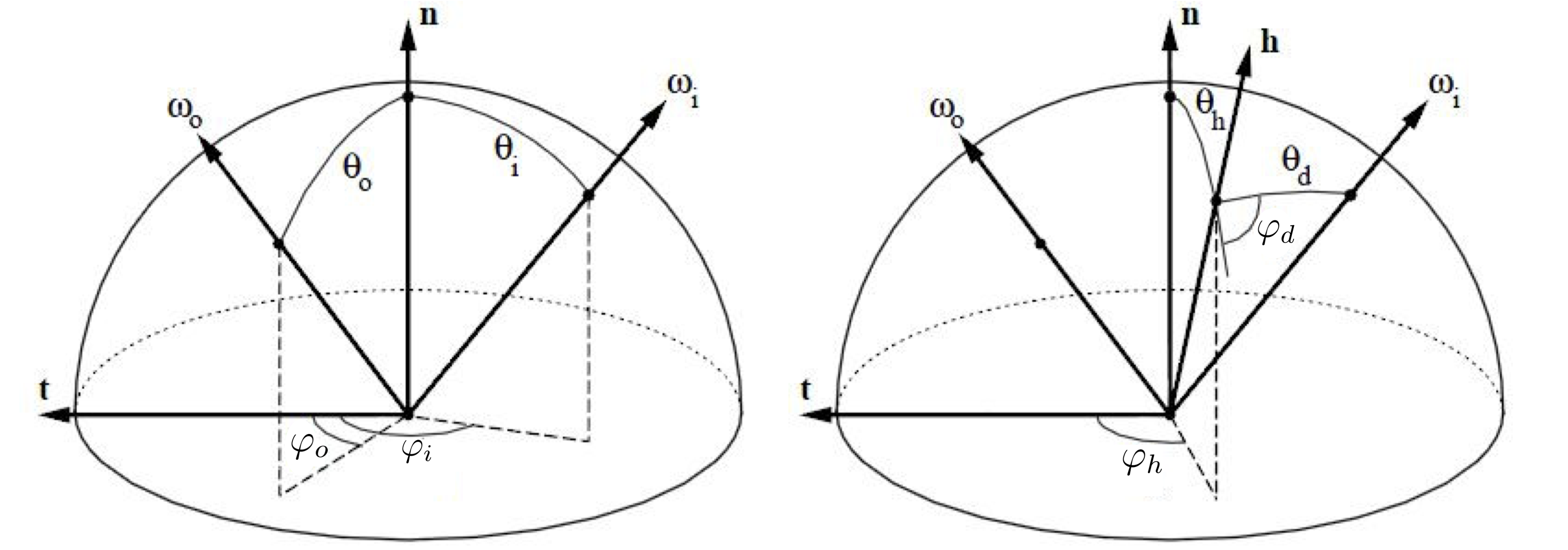}
    \caption{Rusinkiewicz reparameterization of BRDFs.
    $\mathbf{n}$ denotes the surface normal,
    $\mathbf{t}$ the surface tangent,
    $\mathbf{h}$ the half-vector,
    and $\mathbf{w}_i$ and $\mathbf{w}_o$ the incident and outgoing directions.
    Adapted from Rusinkiewicz~\cite{Rusinkiewicz1998}.}
    
    \label{fig:rusinkiewicz}
\end{figure}


Firstly, for our BRDF models (M1 to M5), we first consider the classical BRDF parameterization illustrated on the left of \cref{fig:rusinkiewicz}, where the reflectance is expressed as a function of the incident and outgoing directions $(\mathbf{w}_i, \mathbf{w}_o)$, or equivalently of their associated zenithal and azimuthal angles $(\theta_i, \varphi_i)$ and
$(\theta_o, \varphi_o)$.
In this work, we adopt a simplified angular description commonly used in planetary photometry, by using the phase angle  $\theta_p$ and $\theta_i$.
The incidence angle $\theta_i$ is defined as the zenithal angle between the incident direction and the surface normal $\mathbf{n}$, while the phase angle $\theta_p$ corresponds to the angle between the incident and emission directions.
The phase angle, therefore, implicitly accounts for the emission geometry, such that the
emission angle $\theta_o$ (associated with $\mathbf{w}_o$ in \cref{fig:rusinkiewicz}, left)
does not appear explicitly in the BRDF model. 
Also, the azimuthal angles are not considered in these low-order polynomial formulations to keep a compact parameterization. 
This simplification is justified in our setting, as the camera is most of the time very close to a nadir configuration, making zenithal angles dominant over azimuthal effects.

Then, building upon this classical description, we also investigate a formulation based on the reparameterization introduced by Rusinkiewicz~\cite{Rusinkiewicz1998}, shown on the right of \cref{fig:rusinkiewicz}.
Instead of directly using $(\mathbf{w}_i, \mathbf{w}_o)$, this approach introduces the
half-vector $\mathbf{h} = (\mathbf{w}_i + \mathbf{w}_o) / \lVert \mathbf{w}_i + \mathbf{w}_o \rVert$
and expresses the BRDF as a function of the half-angle $(\theta_h, \varphi_h)$ and a
difference angle $(\theta_d, \varphi_d)$.
This reparameterization provides a clearer interpretation of reflectance behavior along meaningful angular dimensions and is well-suited for analyzing angular reflectance effects beyond the classical incidence–phase formulation.

\section{Additional results on the test dataset}

\begin{figure}
    \centering
    \includegraphics[width=1\linewidth]{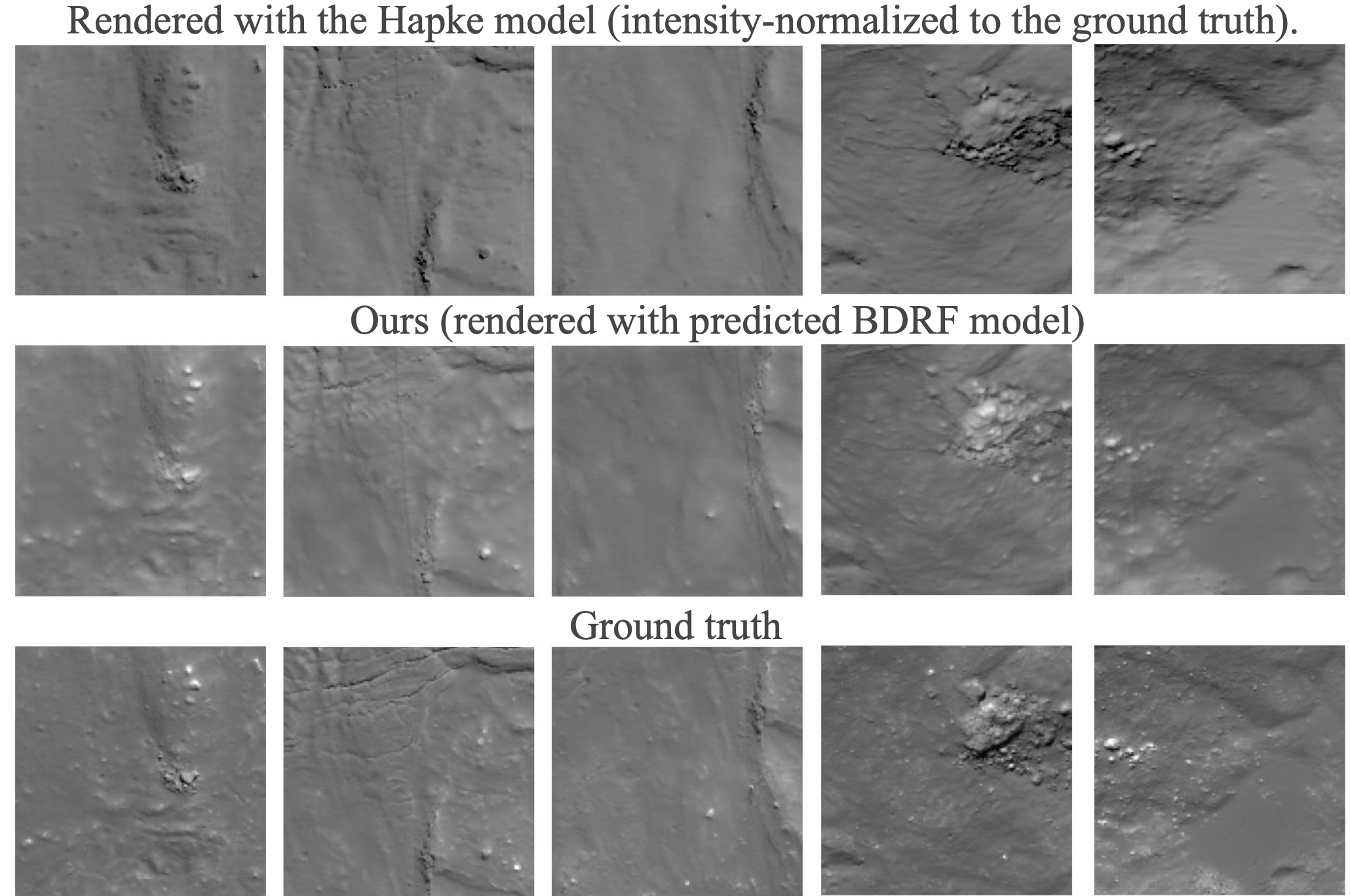}
    \caption{Comparison between normalized Hapke renderings, our predicted BRDF renderings, and ground-truth images. The predicted BRDF better reproduces local shading, brightness variations, and shadow details than the (normalized) Hapke model.}
    \label{fig:res}
\end{figure}
\cref{fig:res} presents a qualitative comparison between images rendered with the classical Hapke reflectance model (top row), those rendered using our predicted spatially varying BRDF (middle row), and the corresponding real lunar observations from the test set (bottom row). 
To enable a fair comparison, the Hapke renderings are intensity-normalized to the dynamic range of the ground-truth images, compensating for the global brightness mismatch inherent to the analytical model.
Even after this normalization, the Hapke images reproduce only the coarse shading tendencies and remain overly smooth, with attenuated contrast and muted highlights.
This limitation is particularly evident in rough or highly sloped areas, where normalization corrects global luminance but cannot restore the local photometric variability that Hapke fails to model.

In contrast, our predicted BRDF produces images whose appearance more closely matches the ground truth, with sharper shading transitions, more realistic brightness variations, and improved recovery of shadow boundaries and fine-scale illumination effects.
These improvements are especially noticeable around small craters and textured slopes, where the learned reflectance better captures the anisotropic photometric response of the lunar regolith under the given illumination geometry.
Overall, these results highlight the benefit of learning spatially varying BRDF parameters directly from DEM-derived geometry, leading to more faithful photometric reconstruction than the (normalized) Hapke model.

\section{Generalization to unseen viewpoints}
\label{sec:unseen_viewpoints}
In these experiments, we evaluate the ability of our model to generalize its BRDF predictions to previously unseen viewpoints over the same underlying DEM. We focus on nadir and oblique views and report qualitative comparisons only, as no ground-truth images are available for these viewpoints.

\subsection{Nadir viewpoint}
\begin{figure}
    \centering
    \includegraphics[width=1\linewidth]{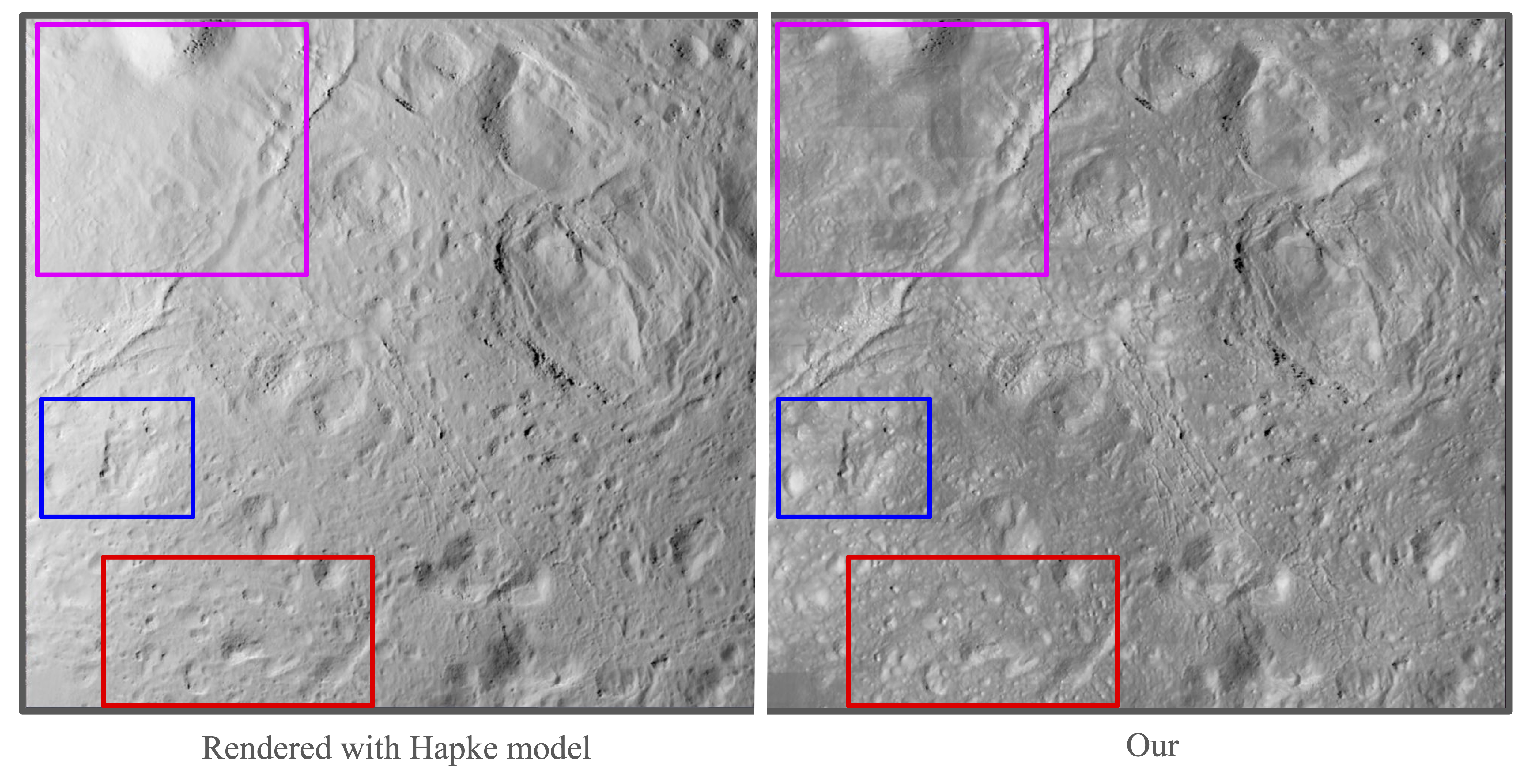}
    \caption{Rendered nadir view of a large area ($\approx \SI{5}{\kilo\meter} \times \SI{5}{\kilo\meter}$) of Tycho crater not seen during training, with former
Hapke model (left) and with the predicted BRDF model (right). We notice in the rendering some squares visible in some areas, which are edge
effects of the BRDF parameters generated (DEM cropping). The colored boxes indicate some areas of interest for viewing improvements in photometric details.}
    \label{fig:unseen-vp-1}
\end{figure}

\begin{figure}
    \centering
    \includegraphics[width=1\linewidth]{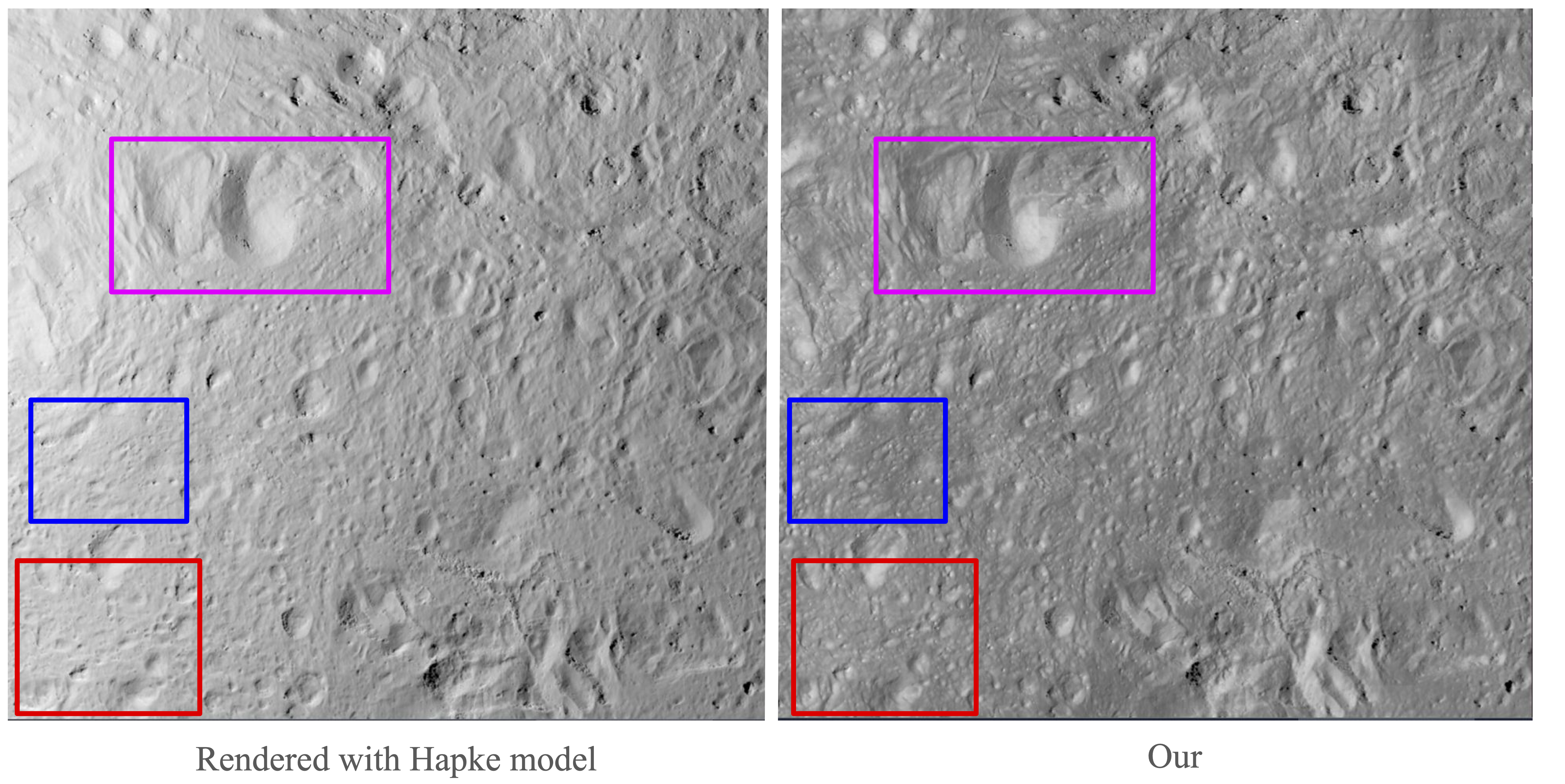}
   \caption{Rendered nadir view of a large area ($\approx \SI{5}{\kilo\meter} \times \SI{5}{\kilo\meter}$) of Tycho crater not seen during training, with former
Hapke model (left) and with the predicted BRDF model (right). We notice in the rendering some squares visible in some areas, which are edge
effects of the BRDF parameters generated (DEM cropping). The colored boxes indicate some areas of interest for viewing improvements in photometric details.}
    \label{fig:unseen-vp-2}
\end{figure}

A nadir viewpoint corresponds to a top-down observation geometry commonly encountered in orbital imaging.
\Cref{fig:unseen-vp-1,fig:unseen-vp-2} present nadir renderings of two large regions
of the Tycho crater, each covering approximately $\SI{5}{\kilo\meter} \times \SI{5}{\kilo\meter}$, which were not observed during training. 

The renderings are generated using the BRDF parameters inferred by our \lunargr\ model.
Clear differences are visible when comparing the former Hapke-based rendering with the predicted BRDF: the latter produces finer photometric details, sharper shading transitions, and spatial reflectance variations that are not captured by the global analytical Hapke model.

These results highlight that lunar surface reflectance is geographically heterogeneous and cannot be accurately represented by a single global BRDF.
Although the network was trained only on local patches extracted from the Tycho crater, it can successfully extrapolate to large-scale, previously unseen nadir views of the same terrain. 
This demonstrates its ability to generalize reflectance properties in scenarios where real observations are spatially incomplete.

\subsection{Oblique viewpoint}
\begin{figure}
    \centering
    \includegraphics[width=0.98\linewidth]{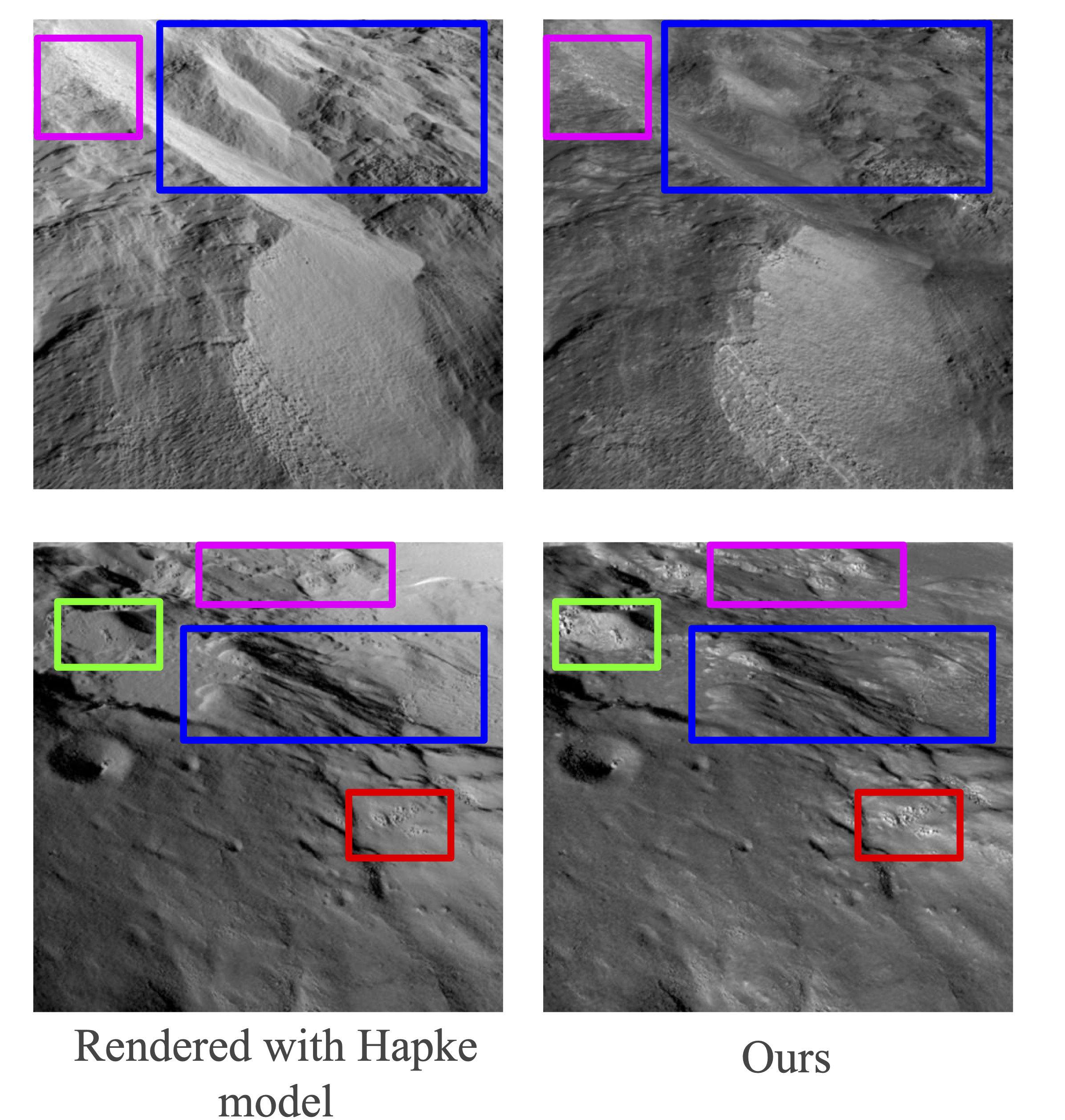}
    \caption{Comparison of lunar surface renderings under an oblique viewpoint.
    \textbf{Top}: rendering using a Hapke BRDF with fixed parameters.
    \textbf{Bottom}: rendering using the BRDF predicted by \lunargr.
    Although no real-world observations are available for this configuration, the comparison highlights the ability of the predicted BRDF to preserve fine-scale photometric variations and local relief contrast under strongly oblique illumination, in contrast to the overly smooth appearance produced by the Hapke model. The colored boxes indicate areas of interest for viewing improvements in photometric details. }
    \label{fig:oblique}
\end{figure}

\Cref{fig:oblique} illustrates a rendering scenario using an oblique viewpoint inspired by the acquisition geometries present in the StereoLunar~\cite{grethen2025adaptingstereovisionobjects} dataset: the cameras, with a field of view of \ang{45}, are tilted, with viewing angles between \ang{20} and \ang{35} and at different altitudes.
Since no real image exists for this configuration, the comparison focuses solely on the differences between the Hapke rendering and the rendering produced with our predicted BRDF model.
The Hapke model produces overly smooth shading and lacks the fine-scale contrast variations that typically emerge under strongly oblique illumination.
Local relief features—such as small crater rims, slope discontinuities, and subtle albedo-driven variations—tend to be flattened or underestimated.
In contrast, the rendering generated using our predicted BRDF exhibits richer photometric structure and sharper shading transitions, which better reflect the spatial variability inferred from the DEM.
Although this viewpoint was never seen during training, the model maintains coherent reflectance patterns across the terrain, demonstrating its ability to generalize to new geometries derived from StereoLunar-like acquisition trajectories.
This experiment highlights the potential of the learned BRDF to extrapolate beyond the training views, even in the absence of corresponding real observations.


\section{Feature-Matching Consistency } 

To further evaluate how well the predicted reflectance preserves the local visual structure of the scene, we measure the consistency of feature correspondences between the ground-truth images and their simulated counterparts. Beyond photometric fidelity, this evaluation aims to assess whether improved reflectance modeling also benefits feature matching performance, which is a key component in vision-based navigation pipelines.
We use the MASt3R \cite{leroy2024grounding} network, specifically the version fine-tuned on moon images~\cite{grethen2025adaptingstereovisionobjects}, to obtain descriptors that are adapted to the photometric and geometric characteristics of Moon terrain. 
MASt3R extracts dense local features, and reciprocal nearest-neighbor matching is performed between the ground-truth image and its simulated counterparts (Hapke-based rendering and learned BRDF). 
Since both images depict the same scene under identical viewing geometry, an accurate reflectance model should produce correspondences that fall at nearly identical pixel locations.
We therefore report the distribution of 2D matching errors as an additional measure of visual similarity. 
\begin{figure}
    \centering
    \includegraphics[width=0.98\linewidth]{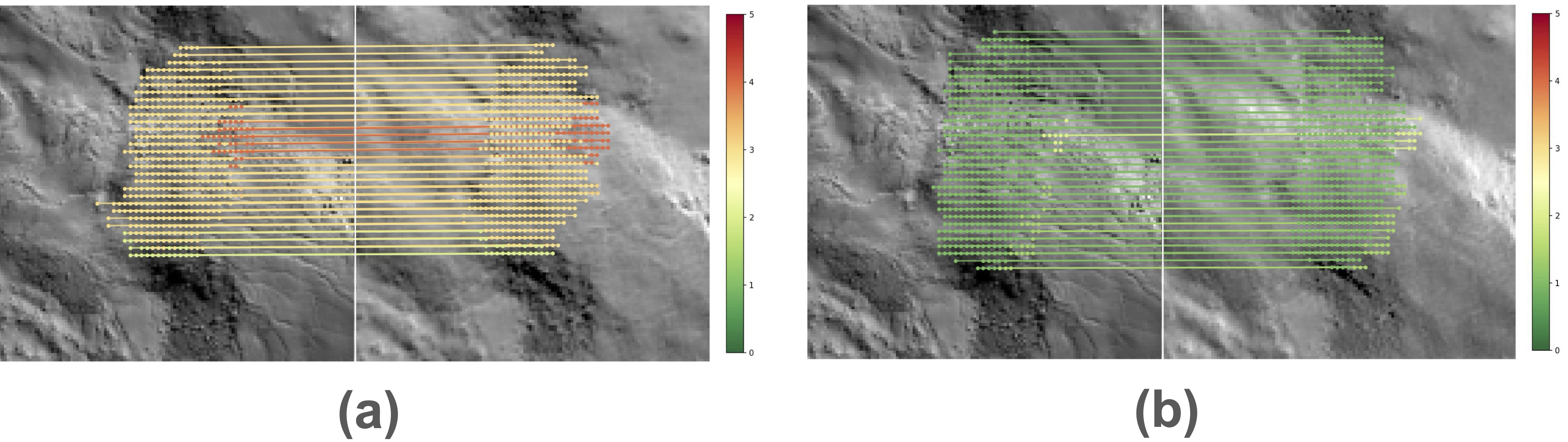}
\caption{Comparison of feature correspondences between simulated and real ground-truth moon images. 
Input images are set to $512 \times 384$ for MASt3R matching; all errors are reported in this resolution.
\textbf{(a):} The normalized Hapke rendering leads to large misalignment (mean error \SI{3.1}{\px}). 
\textbf{(b):} Our learned reflectance yields tight, well-localized matches (mean error \SI{1.0}{\px} over $500$ matches). 
Colors encode per-match error in pixels (dark green = \SI{0}{\px}, dark red = \SI{\geq 5}{\px}).}
\label{fig:result}
\end{figure}

\begin{table}[t]
\centering
\begin{tabular*}{\linewidth}{@{\extracolsep{\fill}}lccc|ccc}
\hline
\textbf{Matcher} 
 & \multicolumn{3}{c|}{\textbf{Hapke}} 
 & \multicolumn{3}{c}{\textbf{Learnt}} \\
\hline
 & $<1$ px ↑ & $<3$ px ↑ & $<5$ px ↑
 & $<1$ px ↑ & $<3$ px ↑ & $<5$ px ↑ \\
\hline
MASt3R 
 & $1.22$ & $36.72$ & $76.62$
 & $\mathbf{8.38}$ & $\mathbf{68.81}$ & $\mathbf{85.26}$ \\

\hline
\end{tabular*}
\caption{Feature-matching consistency for MASt3R fine-tuning for Moon applications) under pixel-distance thresholds.}
\label{tab:matching_thresholds}
\end{table}
The quantitative results in \cref{tab:matching_thresholds} indicate a consistent advantage of the learned reflectance over the Hapke baseline across all pixel-distance thresholds.
Using the MASt3R matcher, the proportion of correspondences within \SI{1}{\px} remains low in absolute terms for both methods, but increases from \SI{1.22}{\percent} with Hapke to \SI{8.38}{\percent} with the learned BRDF. 
More pronounced differences are observed at looser thresholds, with improvements from \SI{36.72}{\percent} to \SI{68.81}{\percent} within \SI{3}{\px}, and from \SI{76.62}{\percent} to \SI{85.26}{\percent} within \SI{5}{\px}. 
These results suggest that the learned SVBRDF yields images that are more structurally consistent with the ground truth, even though exact pixel-level alignment remains limited.
This limitation is to be expected, given that both renderings are generated from the same DEM, which contains geometric inaccuracies that are not present in the ground-truth images.
Nevertheless, as illustrated in \cref{fig:result}, our method produces significantly better-aligned correspondences overall (mean error \SI{1.0}{\px}) than the Hapke baseline (mean error \SI{3.1}{\px}), despite identical geometric constraints.





\clearpage

{
    \scriptsize
    \bibliographystyle{splncs04}    
    \bibliography{strings_abb,main}

@article{Zhang_2021,
   title={{NeRFactor}: neural factorization of shape and reflectance under an unknown illumination},
   volume={40},
   ISSN={1557-7368},
   number={6},
   journal=tog,
   XXpublisher={Association for Computing Machinery (ACM)},
   author={Zhang, Xiuming and Srinivasan, Pratul P. and Deng, Boyang and Debevec, Paul and Freeman, William T. and Barron, Jonathan T.},
   year={2021},
   month=dec, pages={1–18} }

@article{kang2018efficient,
  title={Efficient reflectance capture using an autoencoder.},
  author={Kang, Kaizhang and Chen, Zimin and Wang, Jiaping and Zhou, Kun and Wu, Hongzhi},
  journal={ACM Trans. Graph.},
  volume={37},
  number={4},
  pages={127},
  year={2018}
}

@article{ma2021free,
  title={Free-form scanning of non-planar appearance with neural trace photography},
  author={Ma, Xiaohe and Kang, Kaizhang and Zhu, Ruisheng and Wu, Hongzhi and Zhou, Kun},
  journal=tog,
  volume={40},
  number={4},
  pages={1--13},
  year={2021},
  publisher={ACM New York, NY, USA}
}

@inproceedings{grethen2025adaptingstereovisionobjects,
      title={Adapting Stereo Vision From Objects To 3D Lunar Surface Reconstruction with the StereoLunar Dataset}, 
      author={Clementine Grethen and Simone Gasparini and Geraldine Morin and Jeremy Lebreton and Lucas Marti and Manuel Sanchez-Gestido},
      year={2025},
      booktitle = iccvw,
    pages     = {3751-3760}
}

@article{Ghosh2009,
  title = {Estimating Specular Roughness and Anisotropy from Second Order Spherical Gradient Illumination},
  volume = {28},
  ISSN = {1467-8659},
  number = {4},
  journal = cgf,
  publisher = {Wiley},
  author = {Ghosh,  Abhijeet and Chen,  Tongbo and Peers,  Pieter and Wilson,  Cyrus A. and Debevec,  Paul},
  year = {2009},
  month = jun,
  pages = {1161–1170}
}

@inproceedings{boss2021nerdneuralreflectancedecomposition,
  title         = {NeRD: Neural Reflectance Decomposition from Image Collections},
  author        = {Boss, Mark and Braun, Raphael and Jampani, Varun and Barron, Jonathan T. and Liu, Ce and Lensch, Hendrik P.A.},
  booktitle     = iccv,
  year          = {2021},
}

@article{Deschaintre2018,
  title = {Single-image {SVBRDF} capture with a rendering-aware deep network},
  XXvolume = {37},
  ISSN = {1557-7368},
  XXnumber = {4},
  journal = tog,
  publisher = {Association for Computing Machinery (ACM)},
  author = {Deschaintre,  Valentin and Aittala,  Miika and Durand,  Fredo and Drettakis,  George and Bousseau,  Adrien},
  year = {2018},
  XXmonth = jul,
  XXpages = {1–15}
}

@inproceedings{Hofherr2025Neural,
author = { Hofherr, Florian and Haefner, Bjoern and Cremers, Daniel },
booktitle = wacv,
title = {{ On Neural BRDFs: A Thorough Comparison of State-of-the-Art Approaches }},
year = {2025},
volume = {},
ISSN = {},
pages = {1785-1794},
month =mar}

@article{Kavoosighafi2024,
  title = {Deep SVBRDF Acquisition and Modelling: A Survey},
  XXvolume = {43},
  ISSN = {1467-8659},
  XXnumber = {6},
  journal = cgf,
  XXpublisher = {Wiley},
  author = {Kavoosighafi,  Behnaz and Hajisharif,  Saghi and Miandji,  Ehsan and Baravdish,  Gabriel and Cao,  Wen and Unger,  Jonas},
  year = {2024},
  XXmonth = sep 
}

@ARTICLE{Pieters2016-dm,
  title     = "Space weathering on airless bodies",
  author    = "Pieters, Carle M and Noble, Sarah K",
  journal   = "J. Geophys. Res. Planets",
  publisher = "American Geophysical Union (AGU)",
  volume    =  121,
  number    =  10,
  pages     = "1865--1884",
  month     =  oct,
  year      =  2016,

}

@article{Henriksen2016,
  author = {Henriksen, M. R. and Manheim, M. R. and Speyerer, E. J. and Robinson, M. S. and LROC Team},
  title = {Extracting accurate and precise topography from LROC narrow angle camera stereo observations},
  journal = {Journal of Geophysical Research: Planets},
  volume = {121},
  number = {12},
  pages = {2283--2307},
  year = {2016},
}

@article{Sztrajman2021,
  title = {Neural BRDF Representation and Importance Sampling},
  volume = {40},
  ISSN = {1467-8659},
  number = {6},
  journal = cgf,
  publisher = {Wiley},
  author = {Sztrajman,  Alejandro and Rainer,  Gilles and Ritschel,  Tobias and Weyrich,  Tim},
  year = {2021},
  month = jun,
  pages = {332–346}
}

@article{Guarnera2016,
  title = {BRDF Representation and Acquisition},
  volume = {35},
  ISSN = {1467-8659},
  number = {2},
  journal = cgf,
  publisher = {Wiley},
  author = {Guarnera,  D. and Guarnera,  G.C. and Ghosh,  A. and Denk,  C. and Glencross,  M.},
  year = {2016},
  month = may,
  pages = {625–650}
}

@article{Guo2021Highlight,
author = {Guo, Jie and Lai, Shuichang and Tao, Chengzhi and Cai, Yuelong and Wang, Lei and Guo, Yanwen and Yan, Ling-Qi},
title = {Highlight-aware two-stream network for single-image SVBRDF acquisition},
year = {2021},
issue_date = {August 2021},
publisher = {Association for Computing Machinery},
address = {New York, NY, USA},
XXvolume = {40},
XXnumber = {4},
issn = {0730-0301},
journal = tog,
XXmonth = jul,
articleno = {123},
numpages = {14},
}

@inproceedings{lopes2023materialpaletteextractionmaterials,
    author = {Lopes, Ivan and Pizzati, Fabio and de Charette, Raoul},
    title = {Material Palette: Extraction of Materials from a Single Image},
    booktitle = {CVPR},
    year = {2024},
    project = {https://astra-vision.github.io/MaterialPalette/}
}

@inproceedings{boss2021neuralpilneuralpreintegratedlighting,
  title         = {Neural-PIL: Neural Pre-Integrated Lighting for Reflectance Decomposition},
  author        = {Boss, Mark and Jampani, Varun and Braun, Raphael and Liu, Ce and Barron, Jonathan T. and Lensch, Hendrik P.A.},
  booktitle     = neurips,
  year          = {2021},
}

@article{Nicodemus1965,
  title = {Directional Reflectance and Emissivity of an Opaque Surface},
  volume = {4},
  ISSN = {1539-4522},
  number = {7},
  journal = {Applied Optics},
  publisher = {Optica Publishing Group},
  author = {Nicodemus,  Fred E.},
  year = {1965},
  month = jul,
  pages = {767}
}

@book{Lambert1760,
  author = {Lambert, Johann Heinrich},
  title = {Photometria Sive De Mensura Et Gradibus Luminis, Colorum Et Umbrae},
  publisher = {Eberhard Klett},
  year = {1760},
  address = {Augustae Vindelicorum}
}

@inproceedings{Rusinkiewicz1998,
  title = {A New Change of Variables for Efficient BRDF Representation},
  ISBN = {9783709164532},
  ISSN = {0946-2767},
  booktitle = {Eurographics Workshop on Rendering Techniques},
  author = {Rusinkiewicz,  Szymon M.},
  year = {1998},
  pages = {11–22}
}

@article{Matusik2003,
  title = {A data-driven reflectance model},
  volume = {22},
  ISSN = {1557-7368},
  number = {3},
  journal = tog,
  publisher = {Association for Computing Machinery (ACM)},
  author = {Matusik,  Wojciech and Pfister,  Hanspeter and Brand,  Matt and McMillan,  Leonard},
  year = {2003},
  month = jul,
  pages = {759–769}
}

@inproceedings{Burley2012PhysicallyBasedSA,
  title={Physically-Based Shading at Disney},
  author={Brent Burley},
  booktitle=siggraph,
  year={2012},
}

@article{Dana1999,
author = {Dana, Kristin J. and van Ginneken, Bram and Nayar, Shree K. and Koenderink, Jan J.},
title = {Reflectance and texture of real-world surfaces},
year = {1999},
issue_date = {Jan. 1999},
publisher = {Association for Computing Machinery},
address = {New York, NY, USA},
volume = {18},
number = {1},
journal = {ACM Trans. Graph.},
month = jan,
pages = {1–34},
numpages = {34}
}

@article{dong2010manifold,
  title={Manifold bootstrapping for SVBRDF capture},
  author={Dong, Yue and Wang, Jiaping and Tong, Xin and Snyder, John and Lan, Yanxiang and Ben-Ezra, Moshe and Guo, Baining},
  journal=tog,
  volume={29},
  number={4},
  pages={1--10},
  year={2010},
  publisher={ACM New York, NY, USA}
}

@article{nam2018practical,
  title={Practical {SVBRDF} acquisition of 3d objects with unstructured flash photography},
  author={Nam, Giljoo and Lee, Joo Ho and Gutierrez, Diego and Kim, Min H},
  journal=tog,
  volume={37},
  number={6},
  XXpages={1--12},
  year={2018},
  XXpublisher={ACM New York, NY, USA}
}

@article{Zheng2021NeuralProcessBRDF,
author = {Zheng, Chuankun and Zheng, Ruzhang and Wang, Rui and Zhao, Shuang and Bao, Hujun},
title = {A Compact Representation of Measured BRDFs Using Neural Processes},
year = {2021},
issue_date = {April 2022},
publisher = {Association for Computing Machinery},
address = {New York, NY, USA},
volume = {41},
number = {2},
issn = {0730-0301},
journal = {ACM Trans. Graph.},
month = nov,
articleno = {14},
numpages = {15}
}

@inproceedings{gokbudak2024hypernetworksgeneralizablebrdfrepresentation,
author = {Gokbudak, Fazilet and Sztrajman, Alejandro and Zhou, Chenliang and Zhong, Fangcheng and Mantiuk, Rafal and Oztireli, Cengiz},
title = {Hypernetworks for Generalizable BRDF Representation},
year = {2024},
booktitle = eccv,
pages = {73–89},
}

@inproceedings{leroy2024grounding,
    author = {Leroy, Vincent and Cabon, Yohann and Revaud, Jerome},
    title = {Grounding Image Matching in 3D with {MASt3R}},
    year = {2024},
    XXisbn = {978-3-031-73219-5},
    XXpublisher = {Springer-Verlag},
    XXaddress = {Berlin, Heidelberg},
    XXurl = {https://doi.org/10.1007/978-3-031-73220-1_5},
    XXdoi = {10.1007/978-3-031-73220-1_5},
    booktitle = eccv,
    XXpages = {71–91},
    numpages = {21},
    XXlocation = {Milan, Italy}
}

@book{hapke1993theory,
  author = {Hapke, Bruce},
  title = {{Theory of reflectance and emittance spectroscopy}},
  publisher = {Cambridge University Press},
  year = {1993},
  address = {Cambridge, UK}
}

@inproceedings{mildenhall2020nerf,
 title={NeRF: Representing Scenes as Neural Radiance Fields for View Synthesis},
 author={Ben Mildenhall and Pratul P. Srinivasan and Matthew Tancik and Jonathan T. Barron and Ravi Ramamoorthi and Ren Ng},
 year={2020},
 booktitle={ECCV},
}

@INPROCEEDINGS{Lee2009ClementineBasemap,
       author = {{Lee}, E.~M. and {Gaddis}, L.~R. and {Weller}, L. and {Richie}, J.~O. and {Becker}, T. and {Shinaman}, J. and {Rosiek}, M.~R. and {Archinal}, B.~A.},
        title = "{A New Clementine Basemap of the Moon}",
    booktitle = lpsc,
         year = 2009,
          eid = {2445},
       adsurl = {https://ui.adsabs.harvard.edu/abs/2009LPI....40.2445L}
}

@inproceedings{lebreton2022high,
  title={High performance Lunar landing simulations},
  author={Jérémy Lebreton and Roland Brochard and Nicolas Ollagnier and Matthieu Baudry and Adrien Hadj Salah and Grégory Jonniaux and Keyvan Kanani and Matthieu Le Goff and Aurore Masson},
  booktitle=iac,
  year={2022},
  address={Paris, France},
  month=sep
}

@inproceedings{brochard2018scientific,
      title={Scientific image rendering for space scenes with the {SurRender} software}, 
      author={Roland Brochard and Jérémy Lebreton and Cyril Robin and Keyvan Kanani and Grégory Jonniaux and Aurore Masson and Noela Despré and Ahmad Berjaoui},
      year={2018},
booktitle=esagnc,
      eprint={1810.01423},
      archivePrefix={arXiv},
      primaryClass={astro-ph.IM},
}

@article{Sato2014,
  title = {Resolved Hapke parameter maps of the Moon},
  volume = {119},
  ISSN = {2169-9097},
  XXurl = {http://dx.doi.org/10.1002/2013JE004580},
  XXDOI = {10.1002/2013je004580},
  number = {8},
  journal = planets,
  publisher = {American Geophysical Union (AGU)},
  author = {Sato,  H. and Robinson,  M. S. and Hapke,  B. and Denevi,  B. W. and Boyd,  A. K.},
  year = {2014},
  month = aug,
  pages = {1775–1805}
}

@inproceedings{Parkes2004,
  title = {Planet Surface Simulation with PANGU},
  booktitle = {Space OPS 2004 Conference},
  publisher = {AIAA},
  author = {Parkes,  S.M. and Martin,  I. and Dunstan,  M. and Matthews,  D.},
  year = {2004},
  month = may 
}

@techreport{LRO_Coordinate_System_2008,
  author = {{LRO Project}},
  title = {{A Standardized Lunar Coordinate System for the Lunar Reconnaissance Orbiter (LRO Project White Paper, Version 4)}},
  institution = {{NASA}},
  year = {2008},
  XXmonth = {May},

}

@inproceedings{lrodata,
  author = {NASA},
  title = {Working with Lunar Reconnaissance Orbiter {LROC} Narrow Angle Camera ({NAC}) Data},
  year = {2021}, 
  booktitle = {{LRO} Data Users Workshop},
  address = {Greenbelt, MD, USA}
}

@inproceedings{sarkar2023litnerf,
  title     = {{LitNeRF}: Intrinsic Radiance Decomposition for High-Quality View Synthesis and Relighting of Faces},
  author    = {Kripasindhu Sarkar and Marcel C. Buehler and Gengyan Li and Daoye Wang and Delio Vicini and Jérémy Riviere and Yinda Zhang and Sergio Orts-Escolano and Paulo Gotardo and Thabo Beeler and Abhimitra Meka},
  year      = 2023,
  booktitle = siggraphasia,
  isbn      = {979-8-4007-0315-7/23/12}
}

@article{Mumuni_2024,
   title={A Survey of Synthetic Data Augmentation Methods in Machine Vision},
   volume={21},
   ISSN={2731-5398},
   number={5},
   journal=mir,
   publisher={Springer Science and Business Media LLC},
   author={Mumuni, Alhassan and Mumuni, Fuseini and Gerrar, Nana Kobina},
   year={2024},
   month=mar, pages={831–869} }

@inproceedings{Kajiya1986Rendering,
    author = {Kajiya, James T.},
    title = {The rendering equation},
    year = {1986},
    isbn = {0897911962},
    XXpublisher = {Association for Computing Machinery},
    XXaddress = {New York, NY, USA},
    booktitle = siggraph,
    XXpages = {143–150},
    numpages = {8},
    XXseries = {SIGGRAPH '86}
}

@misc{liu2024lusnaralunarsegmentationnavigation,
      title={LuSNAR:A Lunar Segmentation, Navigation and Reconstruction Dataset based on Muti-sensor for Autonomous Exploration}, 
      author={Jiayi Liu and Qianyu Zhang and Xue Wan and Shengyang Zhang and Yaolin Tian and Haodong Han and Yutao Zhao and Baichuan Liu and Zeyuan Zhao and Xubo Luo},
      year={2024},
      eprint={2407.06512},
      archivePrefix={arXiv},
      primaryClass={cs.CV},
}

@article{Phong1975Illumination,
  title = {Illumination for computer-generated pictures},
  volume = {18},
  ISSN = {1557-7317},
  number = {6},
  journal = acmcom,
  XXpublisher = {Association for Computing Machinery (ACM)},
  author = {Phong,  Bui Tuong},
  year = {1975},
  XXmonth = jun,
  XXpages = {311–317}
}

@article{Ma2023,
  title = {{OpenSVBRDF}: A Database of Measured Spatially-Varying Reflectance},
  volume = {42},
  ISSN = {1557-7368},
  number = {6},
  journal = tog,
  publisher = {Association for Computing Machinery (ACM)},
  author = {Ma,  Xiaohe and Xu,  Xianmin and Zhang,  Leyao and Zhou,  Kun and Wu,  Hongzhi},
  year = {2023},
  month = dec,
  pages = {1–14}
}

@article{Krger2016,
  title = {Geomorphologic mapping of the lunar crater Tycho and its impact melt deposits},
  volume = {273},
  ISSN = {0019-1035},
  journal = {Icarus},
  publisher = {Elsevier BV},
  author = {Kr\"{u}ger,  T. and van der Bogert,  C.H. and Hiesinger,  H.},
  year = {2016},
  month = jul,
  pages = {164–181}
}

@inproceedings{ronneberger2015unetconvolutionalnetworksbiomedical,
  title={U-net: Convolutional networks for biomedical image segmentation},
  author={Ronneberger, Olaf and Fischer, Philipp and Brox, Thomas},
  booktitle=miccai,
  pages={234--241},
  year={2015},
  organization={Springer}
}

@article{Cook1982AReflectance,
    author = {Cook, R. L. and Torrance, K. E.},
    title = {A Reflectance Model for Computer Graphics},
    year = {1982},
    issue_date = {Jan. 1982},
    publisher = {Association for Computing Machinery},
    address = {New York, NY, USA},
    volume = {1},
    number = {1},
    issn = {0730-0301},
    journal = tog,
    month = jan,
    pages = {7–24},
    numpages = {18}
}

@article{Wang2004SSIM,
  author={Zhou Wang and Bovik, A.C. and Sheikh, H.R. and Simoncelli, E.P.},
  journal=tip, 
  title={Image quality assessment: from error visibility to structural similarity}, 
  year={2004},
  volume={13},
  number={4},
  pages={600-612},
}

@INPROCEEDINGS{Zhang2018LPIPS,
  author={Zhang, Richard and Isola, Phillip and Efros, Alexei A. and Shechtman, Eli and Wang, Oliver},
  booktitle=cvpr, 
  title={The Unreasonable Effectiveness of Deep Features as a Perceptual Metric}, 
  year={2018},
  volume={},
  number={},
  pages={586-595},
}

@article{Hapke1986,
  title = {Bidirectional reflectance spectroscopy},
  volume = {67},
  ISSN = {0019-1035},
  XXurl = {http://dx.doi.org/10.1016/0019-1035(86)90108-9},
  XXDOI = {10.1016/0019-1035(86)90108-9},
  number = {2},
  journal = {Icarus},
  XXpublisher = {Elsevier BV},
  author = {Hapke,  Bruce},
  year = {1986},
  month = aug,
  pages = {264–280}
}

@STRING{siggraph = "{ACM SIGGRAPH Conference and Exhibition on Computer Graphics and Interactive techniques}"}

@STRING{cvpr = "{IEEE Conference on Computer Vision and Pattern Recognition}"}

@STRING{eccv = "{European Conference on Computer Vision}"}

@STRING{tog = "{ACM Transactions on Graphics}"}

@STRING{wacv = "{IEEE Winter Conference on Applications of Computer Vision}"}

@STRING{iccv = "{IEEE International Conference on Computer Vision}"}

@STRING{tip = "IEEE Transactions on Image Processing"}

@STRING{siggraphasia = "ACM SIGGRAPH ASIA Course Notes"}

@STRING{siggraph = "ACM SIGGRAPH"}

@STRING{iccvw = "{IEEE International Conference on Computer Vision Workshops}"}

@STRING{acm = {Association for Computing Machinery}}

@STRING{acmcom = {Communications of the ACM}}

@STRING{neurips = {Advances in Neural Information Processing Systems}}

@STRING{esagnc = {International ESA Conference on Guidance, Navigation & Control Systems}}

@string{miccai = {International Conference on Medical Image Computing and Computer-Assisted Intervention}}

@string{mir = {Machine Intelligence Reseach}}

@string{lpsc = {Lunar and Planetary Science Conference} }

@string{iac = {International Astronautical Congress}}

@string{planets = {Journal of Geophysical Research: Planets }}

@STRING{cvpr = "{CVPR}"}

@STRING{eccv = "{ECCV}"}

@STRING{tog = "{ACM Trans. Graph}"}

@STRING{wacv = "{WACV}"}

@STRING{iccv = "{ICCV}"}

@STRING{tip = "Trans. Image Process."}

@STRING{siggraphasia = "SIGGRAPH ASIA"}

@STRING{siggraph = "SIGGRAPH"}

@STRING{iccvw = "ICCVW"}

@STRING{acmcom = {CACM}}

@STRING{neurips = {NeurIPS}}

@STRING{esagnc = {ESA GNC}}

@string{miccai = {MICCAI}}

@string{cgf = {Comput. Graph. Forum}}

@string{mir = {Mach. Intell.}}

@string{lpsc = {LPSC}}

@string{iac = {IAC}}

@string{planets = {JGRE}}
}

\end{document}